%% file: main.tex
\documentclass[10pt,twocolumn,letterpaper]{article}

\usepackage{cvpr}              % To produce the CAMERA-READY version
\usepackage{graphicx}
\usepackage{times}
\usepackage{epsfig}
\usepackage{amsmath}
\usepackage{amssymb}
\usepackage{booktabs}
\usepackage{multirow}
\usepackage{verbatim}
\usepackage{appendix}
\usepackage{subcaption}
\usepackage{adjustbox}
\usepackage{pifont}
\usepackage{enumitem}
\usepackage{makecell}
\usepackage{algorithm, algpseudocode}
\usepackage{listings}
\usepackage[accsupp]{axessibility}  % Improves PDF readability for those with disabilities.
\usepackage{etoolbox}
\makeatletter
\AfterEndEnvironment{algorithm}{\let\@algcomment\relax}
\AtEndEnvironment{algorithm}{\kern2pt\hrule\relax\vskip3pt\@algcomment}
\let\@algcomment\relax
\newcommand\algcomment[1]{\def\@algcomment{\footnotesize#1}}
\renewcommand\fs@ruled{\def\@fs@cfont{\bfseries}\let\@fs@capt\floatc@ruled
  \def\@fs@pre{\hrule height.8pt depth0pt \kern2pt}%
  \def\@fs@post{}%
  \def\@fs@mid{\kern2pt\hrule\kern2pt}%
  \let\@fs@iftopcapt\iftrue}
\makeatother
\newcommand{\trans}[1]{{#1}^{\ensuremath{\mathsf{T}}}} % transpose

\usepackage{comment}
\usepackage[accsupp]{axessibility} % Improves PDF readability for those with disabilities

\usepackage[pagebackref,breaklinks,colorlinks]{hyperref}

\usepackage[capitalize]{cleveref}
\crefname{section}{Sec.}{Secs.}
\Crefname{section}{Section}{Sections}
\Crefname{table}{Table}{Tables}
\crefname{table}{Tab.}{Tabs.}

\def\cvprPaperID{3601} % *** Enter the CVPR Paper ID here
\def\confName{CVPR}
\def\confYear{2023}

\newcommand\blfootnote[1]{%
  \begingroup
  \renewcommand\thefootnote{}\footnote{#1}%
  \addtocounter{footnote}{-1}%
  \endgroup
}

\begin{document}

\title{SCPNet: Semantic Scene Completion on Point Cloud}

\author{\textbf{Zhaoyang Xia}$^{1}$, \textbf{Youquan Liu}$^{1}$, \textbf{Xin Li}$^{1}$, \textbf{Xinge Zhu}$^{2}$, \textbf{Yuexin Ma}$^{3}$,\\ 
\textbf{Yikang Li}$^{1}$, \textbf{Yuenan Hou}$^{1}$$\dagger$, \textbf{and Yu Qiao$^{1}$}\\
$^{1}$Shanghai AI Laboratory $^{2}$The Chinese University of Hong Kong $^{3}$ShanghaiTech University\\
$^{1}$\{houyuenan, liyikang, qiaoyu\}@pjlab.org.cn
}
%\\ $^{2}$zhuxinge123@gmail.com, $^{3}$mayuexin@shanghaitech.edu.cn

\maketitle

\def\algorithmname{SCPNet}

%%%%%%%%% ABSTRACT
\begin{abstract}
\input{sections/00-abstract.tex}
\end{abstract}

%////////////////////////////////
\section{Introduction}
\label{sec:introduction}
%////////////////////////////////
\input{sections/01-introduction.tex}

%////////////////////////////////
\section{Related Work}
\label{sec:relatedwork}
%////////////////////////////////
\input{sections/02-relatedwork.tex}

%////////////////////////////////
\section{Methodology}
\label{sec:methodology}
%////////////////////////////////
\input{sections/03-methodology.tex}

%////////////////////////////////
\section{Experiments}
\label{sec:experiments}
%////////////////////////////////
\input{sections/04-experiment.tex}

\section{Conclusion}\label{conclusion}

To address the challenges of the semantic scene completion task, we propose three solutions from the aspects of the completion network redesign, dense-to-sparse knowledge distillation as well as completion label rectification. The resulting completion network, termed \algorithmname~, achieves superior completion performance in two large-scale semantic scene completion benchmarks, \ie, SemanticKITTI and SemanticPOSS. The learned knowledge on the completion task is also beneficial to the semantic segmentation task.

\noindent \textbf{Acknowledgements.} This work is partially supported by the National Key R\&D Program of China (No.
2022ZD0160100),and in part by  Shanghai Committee of Science and Technology (Grant No. 21DZ1100100).

%\clearpage
%%%%%%%%% REFERENCES
{\small
\bibliographystyle{ieee_fullname}
\bibliography{main}
}

\appendix
\appendixpage
\addappheadtotoc

\begin{table*}[ht]
\caption{Impact of the loss coefficient $\beta$ on the performance.}
\label{tab:loss_coefficient}
\centering
\vskip -0.3cm
%\small{
\begin{adjustbox}{width=\textwidth}
\begin{tabular}{c|c|c|ccccccccccccccccccc}
\hline
$\beta$ & mIoU & completion & \rotatebox{90}{car} & \rotatebox{90}{bicycle} & \rotatebox{90}{motorcycle} & \rotatebox{90}{truck} & \rotatebox{90}{other-vehicle} & \rotatebox{90}{person} & \rotatebox{90}{bicyclist} & \rotatebox{90}{motorcyclist} & \rotatebox{90}{road} & \rotatebox{90}{parking} & \rotatebox{90}{sidewalk} & \rotatebox{90}{other-ground} & \rotatebox{90}{building} & \rotatebox{90}{fence} & \rotatebox{90}{vegetation} & \rotatebox{90}{trunk} & \rotatebox{90}{terrain} & \rotatebox{90}{pole} & \rotatebox{90}{traffic-sign} \\
\hline
\hline
4000 & 35.1 & 49.9 & 49.5 & 25.4 & 28.9 & 47.0 & 40.2 & 15.3 & 16.1 & \textbf{5.1} & 70.2 & 58.5 & 51.4 & 11.5 & 33.0 & 30.1 & 40.3 & 31.5 & 49.3 & 37.0 & 27.5 \\
3000 & \textbf{37.2} & 49.9 & \textbf{50.5} & \textbf{28.5} & \textbf{31.7} & \textbf{58.4} & 41.4 & \textbf{19.4} & \textbf{19.9} & 0.2 & \textbf{70.5} & \textbf{60.9} & \textbf{52.0} & \textbf{20.2} & \textbf{34.1} & \textbf{33.0} & 35.3 & \textbf{33.7} & \textbf{51.9} & \textbf{38.3} & 27.5 \\
2000 & 35.3 & \textbf{50.4} & 50.0 & 25.8 & 31.3 & 56.4 & \textbf{41.6} & 17.2 & 9.8 & 0.0 & 70.0 & 58.0 & 51.3 & 8.3 & 31.7 & 28.7 & \textbf{40.7} & 32.7 & 51.6 & 38.1 & \textbf{27.9} \\
1000 & 35.0 & 48.8 & 49.2 & 27.1 & 29.6 & 56.2 & 36.4 & 16.4 & 14.2 & 0.0 & 69.6 & 57.7 & 50.7 & 7.7 & 29.7 & 30.4 & 34.7 & 30.3 & 48.0 & 37.0 & 27.4 \\
\hline
\end{tabular}
\end{adjustbox}
\vspace{-2ex}
\end{table*}

\begin{table*}[ht]
\caption{Impact of DSKD loss on the performance.}
\label{tab:dskd_loss}
\centering
\vskip -0.3cm
%\small{
\begin{adjustbox}{width=\textwidth}
\begin{tabular}{c|c|c|ccccccccccccccccccc}
\hline
Methods & mIoU & completion & \rotatebox{90}{car} & \rotatebox{90}{bicycle} & \rotatebox{90}{motorcycle} & \rotatebox{90}{truck} & \rotatebox{90}{other-vehicle} & \rotatebox{90}{person} & \rotatebox{90}{bicyclist} & \rotatebox{90}{motorcyclist} & \rotatebox{90}{road} & \rotatebox{90}{parking} & \rotatebox{90}{sidewalk} & \rotatebox{90}{other-ground} & \rotatebox{90}{building} & \rotatebox{90}{fence} & \rotatebox{90}{vegetation} & \rotatebox{90}{trunk} & \rotatebox{90}{terrain} & \rotatebox{90}{pole} & \rotatebox{90}{traffic-sign} \\
\hline
\hline
SCPNet w/ DSKD & \textbf{37.2} & \textbf{49.9} & \textbf{50.5} & \textbf{28.5} & \textbf{31.7} & \textbf{58.4} & 41.4 & \textbf{19.4} & \textbf{19.9} & \textbf{0.2} & \textbf{70.5} & \textbf{60.9} & \textbf{52.0} & \textbf{20.2} & \textbf{34.1} & \textbf{33.0} & 35.3 & \textbf{33.7} & \textbf{51.9} & \textbf{38.3} & \textbf{27.5} \\
SCPNet w/o DSKD & 34.4 & 48.5 & 48.5 & 26.4 & 28.1 & 54.6 & \textbf{41.7} & 14.5 & 13.1 & 0.0 & 70.2 & 58.3 & 51.3 & 2.9 & 31.7 & 30.4 & \textbf{37.9} & 31.6 & 49.2 & 36.7 & 25.7 \\
\hline
\end{tabular}
\end{adjustbox}
\vspace{-2ex}
\end{table*}

\begin{table*}[!t]
\caption{Impact of downsampling on the performance.}
\label{tab:downsampling}
\centering
\vskip -0.3cm
%\small{
\begin{adjustbox}{width=\textwidth}
\begin{tabular}{c|c|c|ccccccccccccccccccc}
\hline
Methods & mIoU & completion & \rotatebox{90}{car} & \rotatebox{90}{bicycle} & \rotatebox{90}{motorcycle} & \rotatebox{90}{truck} & \rotatebox{90}{other-vehicle} & \rotatebox{90}{person} & \rotatebox{90}{bicyclist} & \rotatebox{90}{motorcyclist} & \rotatebox{90}{road} & \rotatebox{90}{parking} & \rotatebox{90}{sidewalk} & \rotatebox{90}{other-ground} & \rotatebox{90}{building} & \rotatebox{90}{fence} & \rotatebox{90}{vegetation} & \rotatebox{90}{trunk} & \rotatebox{90}{terrain} & \rotatebox{90}{pole} & \rotatebox{90}{traffic-sign} \\
\hline
\hline
w/o downsampling & \textbf{37.2} & 49.9 & \textbf{50.5} & \textbf{28.5} & \textbf{31.7} & \textbf{58.4} & \textbf{41.4} & \textbf{19.4} & \textbf{19.9} & \textbf{0.2} & 70.5 & \textbf{60.9} & \textbf{52.0} & \textbf{20.2} & \textbf{34.1} & \textbf{33.0} & 35.3 & \textbf{33.7} & \textbf{51.9} & \textbf{38.3} & \textbf{27.5} \\
w/ downsampling & 33.1 & \textbf{51.0} & 48.7 & 21.3 & 28.9 & 40.3 & 30.3 & 17.6 & 16.0 & 0.0 & \textbf{70.7} & 58.8 & 51.3 & 11.5 & 33.6 & 29.4 & \textbf{41.2} & 32.3 & 51.5 & 35.9 & 8.9 \\
\hline
\end{tabular}
\end{adjustbox}
\vspace{-1ex}
\end{table*}

\section{Ablation studies}

\noindent \textbf{Loss coefficient $\beta$.} We investigate the effect of the loss coefficient of the DSKD loss on the final performance. As shown in Table~\ref{tab:loss_coefficient}, when we change the loss coefficient from 1, 000 to 4, 000, the completion performance of SCPNet first improves and then declines. Therefore, we set the loss coefficient of the DSKD loss as 3, 000 to obtain the best performance.

\noindent \textbf{Detailed performance comparison on DSKD loss.} The detailed performance comparison of SCPNet with and without DSKD is shown in Table ~\ref{tab:dskd_loss}. On motorcycle, truck, person and bicyclist, the proposed DSKD loss can bring more than 3 IoU improvement.

\noindent \textbf{Detailed performance comparison on the downsampling operation.} We examine the effect of adding the downsampling operation to the completion sub-network of SCPNet. The detailed performance comparison of SCPNet with and without the downsampling operation is shown in Table ~\ref{tab:downsampling}. It is apparent that the completion performance drops significantly, especially for truck, other-vehicle, other-ground and traffic-sign. The severe performance degradation strongly shows the necessity of removing the lossy downsampling operation for the completion sub-network.

\section{Elaborated implementation details}

\noindent \textbf{Range mismatch.} On SemanticKITTI, the point cloud range used by our segmentation sub-network, \ie, Cylinder3D, is [-36.2, 36.2] m, [-36.2, 36.2] m and [-4, 2] m for x, y, z, respectively. For semantic scene completion, the range of the completion labels is [0, 51.2] m, [-25.6, 25.6] m and [-2, 4.4] m for x, y, z, respectively. The range mismatch problem will cause the existence of many empty voxels, which will significantly hamper the completion performance. To address this problem, we directly use the point cloud range of the completion labels. %take the intersection of two ranges and the range for the completion task is set as [0, 36.2] m, [-25.6, 25.6] m and [-2, 2] m for x, y and z, respectively.

\noindent \textbf{Why conv bias and BN layers breaks the sparsity of voxel features.} The voxel features, which are treated as the input of the completion sub-network, are sparse, \ie, only a part of the whole voxel space is occupied. The completion sub-network uses the vanilla dense convolution for dilation. However, the bias of 3D convolution weight, the mean and variance of the Batch Normalization (BN) layers will result in non-zero values of all empty voxel positions. This will cause all empty voxel features to become occupied, which breaks the sparsity of the original voxel features and significantly increases the computation burden of the segmentation sub-network.

\noindent \textbf{How does changing the random seed influence the mIoU values.} We conduct experiments on SemanticKITTI using three different random seeds, \ie, 100, 240 and 666. Experiments on SemanticKITTI show that the performance variance of \algorithmname~is within 0.3 mIoU.

\noindent \textbf{Apply the proposed distillation loss to other architectures.} We apply the DSKD loss to JS3CNet. It improves the performance of JS3CNet from 24.0 mIoU to 26.2 mIoU on SemanticKITTI val set.

\noindent \textbf{Apply label rectification to other models.} We apply label rectification to JS3CNet. Experimental results show that on SemanticKITTI val set, the proposed label rectification can bring considerable gains to JS3CNet, improving the performance from 24.0 mIoU to 26.8 mIoU.

\noindent \textbf{Computational impact of the proposed adjustments.} We calculate the computational overhead of the completion sub-network and finds that it only introduces around 24.4 ms overhead.

\noindent \textbf{Error bands for the results.} We run the experiments on SemanticKITTI and SemanticPOSS datasets for three times. The performance variance of SCPNet on these benchmarks is within 0.3 mIoU.

\noindent \textbf{Why panoptic labels are useful in label rectification.} The panoptic labels provide instance-level annotations for those thing classes (\eg, cars and persons) and these instance-level annotations are helpful to remove the long traces of moving objects in completion labels which only provide semantic segmentation annotations and do not differentiate each single instance.

\begin{table}[!t]
\caption{Training and inference time using A100.}
\label{tab:train_infer}
\centering
\vskip -0.1cm
%\small{
\begin{tabular}{l|c|c|c}
\hline
Methods & Train (h) & Inference (ms) & mIoU \\ %& Completion \\
\hline
\algorithmname & 34 & 143.2 & 37.2 \\ %& 48.5 \\
JS3CNet & 28 & 120.6 & 24.0 \\
\hline
\end{tabular}
\vspace{-2ex}
\end{table}

\noindent \textbf{Training and inference time and a comparison with SOTA.} We summarized the training and inference time between JS3CNet and our \algorithmname~in Table~\ref{tab:train_infer}. Our SCPNet has comparable training and inference time but exhibits much better completion performance than JS3CNet.

\section{Qualitative results}

\begin{figure*}[t]
 \centering
 \includegraphics[width=1.0\linewidth]{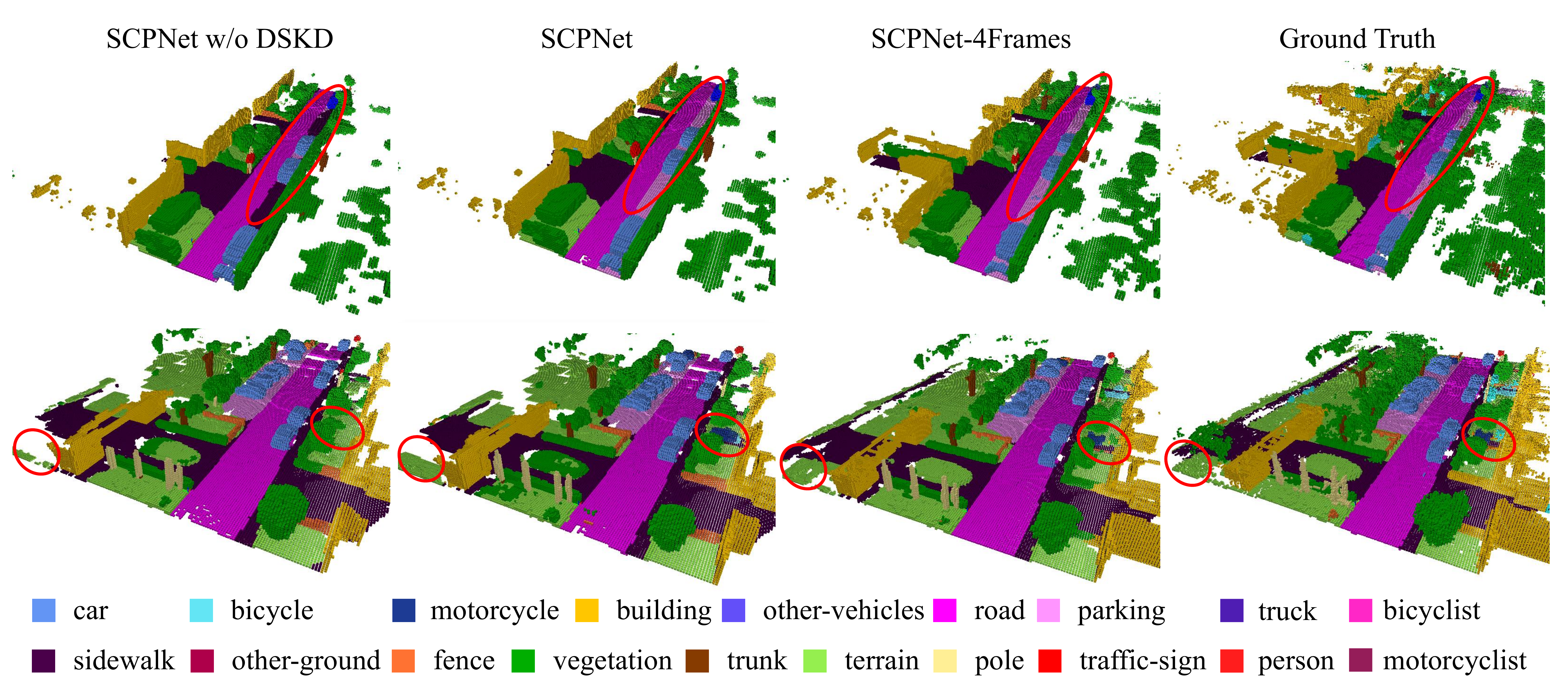}
 \vskip -0.3cm
 \caption{Visual comparison of \algorithmname~with and without DSKD on the SemanticKITTI~\cite{behley2019semantickitti} validation set. From left to right: \algorithmname~without the DSKD loss, \algorithmname~with DSKD, \algorithmname-4Frames and ground-truth. Different color represents different class.}
 \centering
 \vskip -0.1cm
 \label{fig:visual_compare_dskd}
\end{figure*}

We provide visual comparison of \algorithmname~with and without DSKD in Fig.~\ref{fig:visual_compare_dskd}. Compared with SCPNet without DSKD, the single-frame SCPNet with DSKD achieves better completion and segmentation performance by distilling dense and relation-based information from the multi-frame teacher model. SCPNet without DSKD performs badly on the parking areas and small objects while SCPNet with DSKD exhibits much better completion performance owing to the proposed distillation objective.

\begin{figure*}[t]
 \centering
 \includegraphics[width=1.0\linewidth]{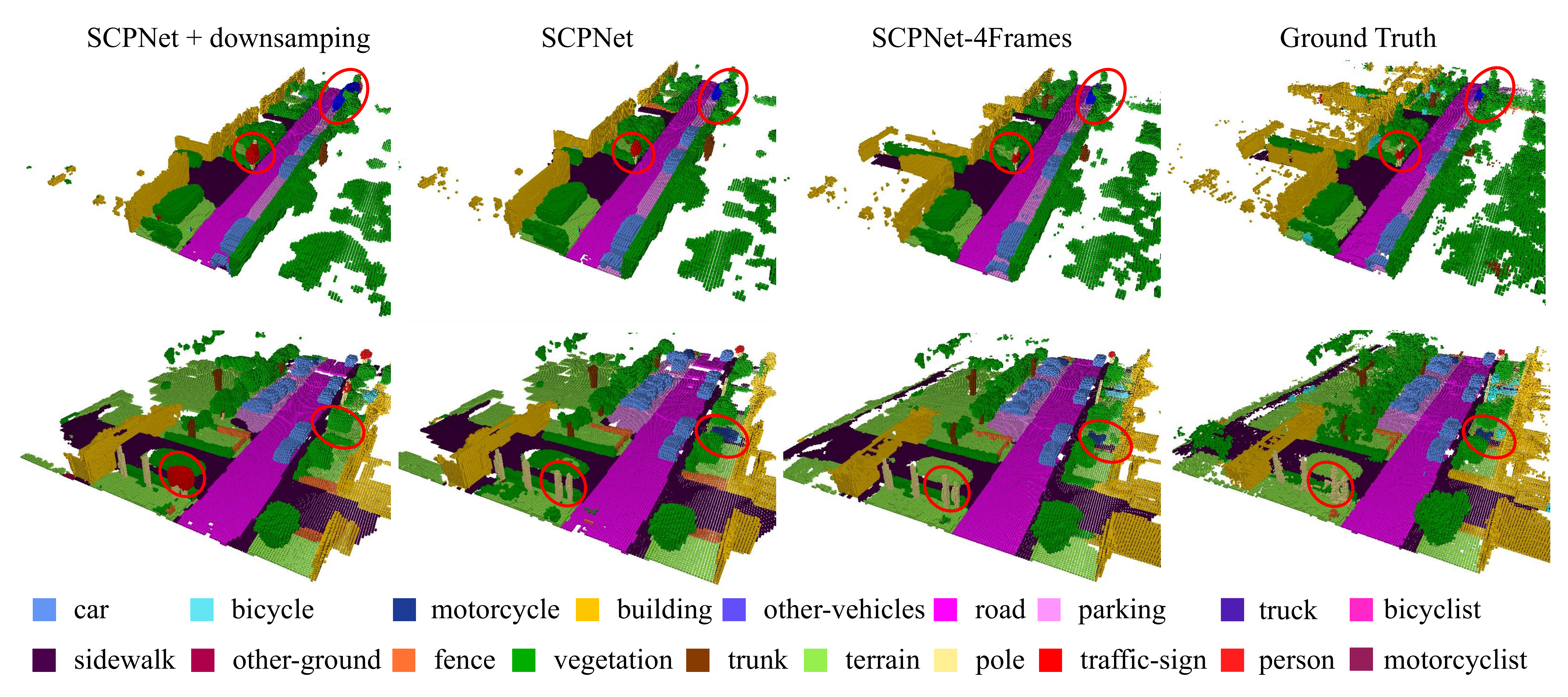}
 \vskip -0.1cm
 \caption{Visual comparison of \algorithmname~with and without the downsampling operation on the SemanticKITTI~\cite{behley2019semantickitti} validation set. From left to right: \algorithmname~with downsampling operation, \algorithmname~without downsampling operation, \algorithmname-4Frames and ground-truth. Different color represents different class.}
 \centering
 \vskip -0.1cm
 \label{fig:visual_compare_downsampling}
\end{figure*}

And we also provide visual comparison between original \algorithmname~and \algorithmname~with downsampling and upsampling operations. As can be seen from Fig.~\ref{fig:visual_compare_downsampling}, the downsampling and upsampling operations will cause over dilation and shape distortion for these objects marked by the red ellipses.

\end{document}

%% file: sections/00-abstract.tex
% !TEX root = ../main.tex

Training deep models for semantic scene completion (SSC) is challenging due to the sparse and incomplete input, a large quantity of objects of diverse scales as well as the inherent label noise for moving objects. To address the above-mentioned problems, we propose the following three solutions: 1) \textbf{Redesigning the completion sub-network.} We design a novel completion sub-network, which consists of several Multi-Path Blocks (MPBs) to aggregate multi-scale features and is free from the lossy downsampling operations. 2) \textbf{Distilling rich knowledge from the multi-frame model.} We design a novel knowledge distillation objective, dubbed Dense-to-Sparse Knowledge Distillation (DSKD). It transfers the dense, relation-based semantic knowledge from the multi-frame teacher to the single-frame student, significantly improving the representation learning of the single-frame model. 3) \textbf{Completion label rectification.} We propose a simple yet effective label rectification strategy, which uses off-the-shelf panoptic segmentation labels to remove the traces of dynamic objects in completion labels, greatly improving the performance of deep models especially for those moving objects. Extensive experiments are conducted in two public SSC benchmarks, \ie, SemanticKITTI and SemanticPOSS. Our \algorithmname~ranks \textbf{1}st on SemanticKITTI semantic scene completion challenge and surpasses the competitive S3CNet~\cite{s3cnet} by \textbf{7.2} mIoU. \algorithmname~also outperforms previous completion algorithms on the SemanticPOSS dataset. Besides, our method also achieves competitive results on SemanticKITTI semantic segmentation tasks, showing that knowledge learned in the scene completion is beneficial to the segmentation task. %Our code will be available at \url{https://github.com/cardwing/Codes-for-SCPNet}.
%since both geometry and semantics of the scene need to be inferred from the input incomplete and sparse view

%% file: sections/01-introduction.tex
% !TEX root = ../main.tex

\blfootnote{$\dagger$: Corresponding author.}

\begin{figure*}[t]
 \centering
 \includegraphics[width=1.0\linewidth]{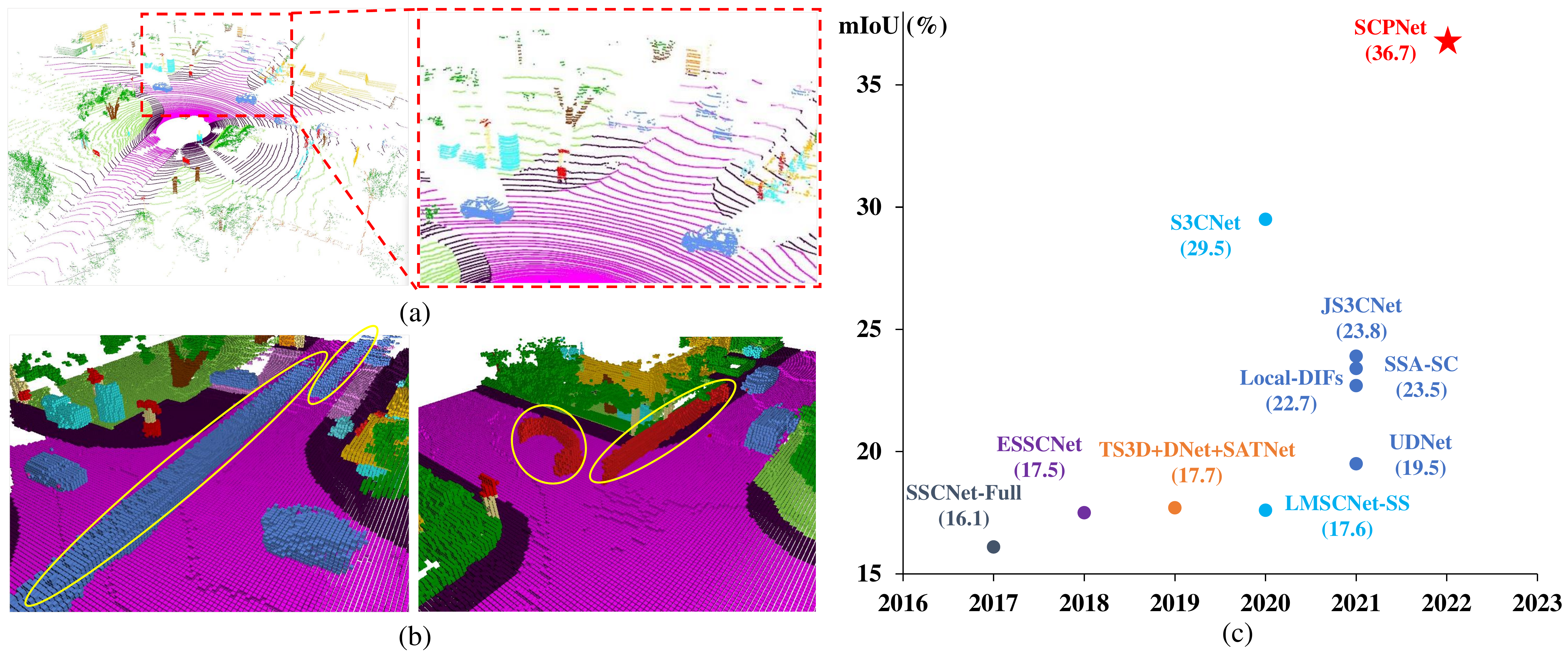}
 \vskip -0.3cm
 \caption{Left: challenges of semantic scene completion, \ie, (a) sparse and incomplete input, varying completion difficulty for objects of diverse scales and (b) long traces of dynamic objects in completion labels. Traces of cars and persons are highlighted by yellow ellipses. Right: (c) performance comparison between competitive completion algorithms and \algorithmname~on SemanticKITTI semantic scene completion challenge.}
 \centering
 \vskip -0.6cm
 \label{fig:motivation}
\end{figure*}

Semantic scene completion (SSC)~\cite{completion_survey} aims at inferring both geometry and semantics of the scene from an incomplete and sparse observation, which is a crucial component in 3D scene understanding. Performing semantic scene completion in the outdoor scenarios is challenging due to the sparse and incomplete input, a large quantity of objects of diverse scales as well as the inherent label noise for those moving objects (See Fig.~\ref{fig:motivation} (a)).

Recent years have witnessed an explosion of methods in the outdoor scene completion field~\cite{s3cnet,js3cnet,ssa-sc,sscnet,wilson2022motionsc,lmscnet}. For example, S3CNet~\cite{s3cnet} performs 2D and 3D completion tasks jointly and achieves impressive performance on the SemanticKITTI leaderboard~\cite{behley2019semantickitti}. JS3C-Net~\cite{js3cnet} performs semantic segmentation first and then feeds the segmentation features to the completion sub-network. The coarse-to-fine refinement module is further put forward to improve the completion quality. Although significant progress has been achieved in this area, these methods heavily rely on the voxelwise completion labels and show unsatisfactory completion performance on the small, distant objects and crowded scenes. Moreover, the long traces of the dynamic objects in original completion labels will hamper the learning of the completion models, which is overlooked in the previous literature~\cite{completion_survey}.

To address the preceding problems, we propose three solutions from the aspects of the completion sub-network redesign, distillation of multi-frame knowledge as well as completion label rectification. Specifically, we first make a comprehensive overhaul of the completion sub-network. We adopt the completion-first principle and make the completion module directly process the raw voxel features. Besides, we avoid the use of downsampling operations since they inevitably introduce information loss and cause severe misclassification for those small objects and crowded scenes. To improve the completion quality on objects of diverse scales, we design Multi-Path Blocks (MPBs) with varied kernel sizes, which aggregate multi-scale features and fully utilize the rich contextual information.

Second, to combat against the sparse and incomplete input signals, we make the single-scan student model distil knowledge from the multi-frame teacher model. However, mimicking the probabilistic knowledge of each point/voxel brings marginal gains. Instead, we propose to distill the pairwise similarity information. Considering the sparsity and unorderness of features, we align the features using their indices and then force the consistency between the pairwise similarity maps of student features and those of teacher features, make the student benefit from the relational knowledge of the teacher. The resulting Dense-to-Sparse Knowledge Distillation objective is termed DSKD, which is specifically designed for the scene completion task.

Finally, to address the long traces of those dynamic objects in the completion labels, we propose a simple yet effective label rectification strategy. The core idea is to use off-the-shelf panoptic segmentation labels to remove the traces of dynamic objects in completion labels. The rectified completion labels are more accurate and reliable, greatly improving the completion qualities of deep models on those moving objects.

%2) Multi-frame to single-frame knowledge distillation. The first to apply knowledge distillation to semantic scene completion.
%3) Present a simple yet effective solution to handle the smear problem. Correction of current completion labels for those moving objects. 

We conduct experiments on two large-scale outdoor scene completion benchmarks, \ie, SemanticKITTI~\cite{behley2019semantickitti} and SemanticPOSS~\cite{semanticposs}. Our \algorithmname~ranks \textbf{1}st on SemanticKITTI semantic scene completion challenge~\footnote{https://codalab.lisn.upsaclay.fr/competitions/7170\#results till 2022-11-12 00:00 Pacific Time, and our method is termed \algorithmname.} and outperforms the S3CNet~\cite{s3cnet} by \textbf{7.2} mIoU. \algorithmname~also achieves better performance than other completion algorithms on the SemanticPOSS dataset. The learned knowledge from the completion task also benefits the segmentation task, making our \algorithmname~achieve superior performance on the SemanticKITTI semantic segmentation task.

Our contributions are summarized as follows.
\begin{itemize}

\item {The comprehensive redesign of the completion sub-network. We unveil several key factors to building strong completion networks.}
\item {To cope with the sparsity and incompleteness of the input, we propose to distill the dense relation-based knowledge from the multi-frame model. Note that we are the \textbf{first} to apply knowledge distillation to the semantic scene completion task.}
\item {To address the long traces of moving objects in completion labels, we present the completion label rectification strategy.}
\item {Our \algorithmname~ranks 1st on SemanticKITTI semantic scene completion challenge, outperforming the previous SOTA S3CNet~\cite{s3cnet} by \textbf{7.2} mIoU. Competitive performance is also shown in SemanticPOSS completion task and SemanticKITTI semantic segmentation task.}
\end{itemize}

%% file: sections/02-relatedwork.tex
% !TEX root = ../main.tex
%\noindent \textbf{Indoor semantic scene completion.}
%PoinTR, GRNet

%SSCNet is the earliest network designed for semantic scene completion.

\begin{figure*}[!ht]
 \centering
 \includegraphics[width=1.0\linewidth]{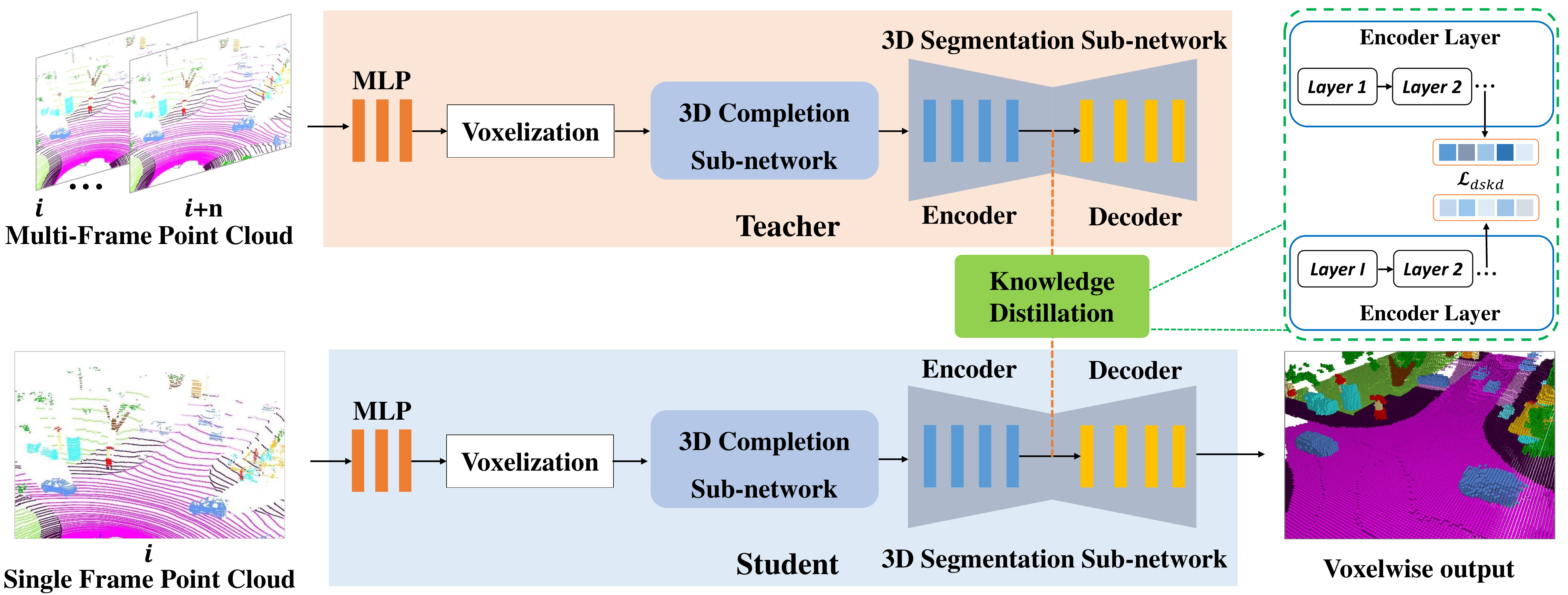}
 \vskip -0.3cm
 \caption{Framework overview of \algorithmname. There are two SCPNets, one is teacher and the other is student, and they have the same architecture. The teacher takes multi-frame point cloud as input while the student takes the single-frame point cloud as input. The dense-to-sparse knowledge distillation loss is proposed to transfer the dense semantic knowledge from teacher to student. In each \algorithmname, there are two sub-networks, \ie, the completion sub-network and the segmentation sub-network. The point cloud is first processed by a stack of MLPs to extract point features. These point features are voxelized and then fed to the completion sub-network to produce denser voxel features. The produced voxel features are further fed to the segmentation sub-network to generate the ultimate voxelwise output. For the completion sub-network, it is comprised of several multi-path blocks, free from the lossy downsampling operations. For the segmentation sub-network, it is adapted from the Cylinder3D network.}
 \centering
 \vskip -0.4cm
 \label{fig:framework}
\end{figure*}

\noindent \textbf{Semantic scene completion.} Early works on scene completion mainly concentrates on the indoor scenarios~\cite{survey_point_cloud_completion,sscnet,scancomplete,satnet}. The point cloud in indoor scenarios is dense, small-scale and has uniform density. By contrast, point cloud in outdoor scenes is sparse, large-scale and has varying density, which poses great challenges to the semantic scene completion algorithms~\cite{completion_survey,s3cnet,js3cnet,ssa-sc,wilson2022motionsc}. Various algorithms have been proposed, for instance, LMSCNet~\cite{lmscnet} uses a mixture of 2D and 3D convolutions to build the efficient completion backbone. S3CNet~\cite{s3cnet} performs 2D and 3D scene completion simultaneously, and fuses the results by the proposed dynamic voxel fusion module. JS3C-Net~\cite{js3cnet} attaches the completion network to the segmentation backbone and refines the completion outputs via the point-voxel interaction module. Compared with previous completion networks, our \algorithmname~is free from the lossy downsampling operations. Besides, our network is built on several MPBs that aggregate multi-scale features and can achieve high completion quality in objects of various sizes. 

%Semantic Segmentation-assisted Scene Completion for LiDAR Point Clouds, IROS 2021

%MotionSC: Data Set and Network for Real-Time Semantic Mapping in Dynamic Environments, RA-L 2022

\noindent \textbf{Knowledge distillation.} Knowledge distillation (KD) originates from the pioneering work of G. Hinton~\etal~\cite{hinton2015distilling}. Its primary objective is to transfer the dark knowledge from the large over-parameterized teacher model to the small compact student model. A large number of methods have been proposed and various forms of knowledge~\cite{romero2015fitnets,zagoruyko2016paying,tung2019similarity,xu2022mind} have been designed, \eg, intermediate features~\cite{romero2015fitnets,fmnet}, visual attention maps~\cite{zagoruyko2016paying,sad}, region-level affinity scores~\cite{hou2020inter}, similarity scores of different samples~\cite{tung2019similarity,xing2021categorical}, \etc. It is noteworthy that the majority of the distillation methods concentrate on the 2D tasks. Only a few distillation methods have focused on 3D domains, \eg, PVKD~\cite{pvkd2022} and SparseKD~\cite{sparsekd}. To our knowledge, this is the first work that applies knowledge distillation to the semantic scene completion task. We propose to transfer the dense, relation-based semantic knowledge from the multi-frame model to the single-frame one. %The most relevant work to ours is~\cite{pvkd2022}. However, there are several distinctions. First, \cite{wang2020multi} focuses on 3D detection while ours is concentrated on semantic scene completion. Second, \cite{wang2020multi} aims to distil element-wise feature knowledge while our method distils the relation-based structural knowledge from the multi-frame teacher model. %Til now, we are the first to apply knowledge distillation to semantic scene completion. A specifically designed KD objective is proposed to distil the dense semantic knowledge from the multi-frame teacher model.

%% file: sections/03-methodology.tex
% !TEX root = ../main.tex

The objective of the semantic scene completion task is to infer the complete geometric and semantic layout given the incomplete and sparse input. Formally, given the input point cloud $\mathbf{X} \in \mathbb{R}^{N \times 3}$, the network needs to assign a label to each voxel of the $L \times W \times H$ voxel space to indicate whether it is empty or belongs to a specific class $c \in \{0, 1, 2,..., C-1 \}$, where $C$ is the number of classes, $L$, $W$ and $H$ are the length, width and height of the voxel space, respectively. We denote the voxelwise output as $\mathbf{O} \in \mathbb{R}^{L \times W \times H \times C}$.

\subsection{Framework overview}

As shown in Fig.~\ref{fig:framework}, our \algorithmname~is comprised of two sub-networks, \ie, the completion sub-network and the segmentation sub-network. The completion sub-network is designed based on several key design principles. The segmentation sub-network is built upon Cylinder3D~\cite{zhu2021cylindrical,zhu2021cylindrical-tpami}, with some minor modifications. In the following sections, we first detail the completion sub-network and introduce several design principles that are vital to building a strong completion sub-network. The knowledge distillation of multi-frame model and the completion label rectification will be explained thereafter.

\subsection{Redesigning the Completion Sub-network}

\begin{figure}[!ht]
 \centering
 \includegraphics[width=1.0\linewidth]{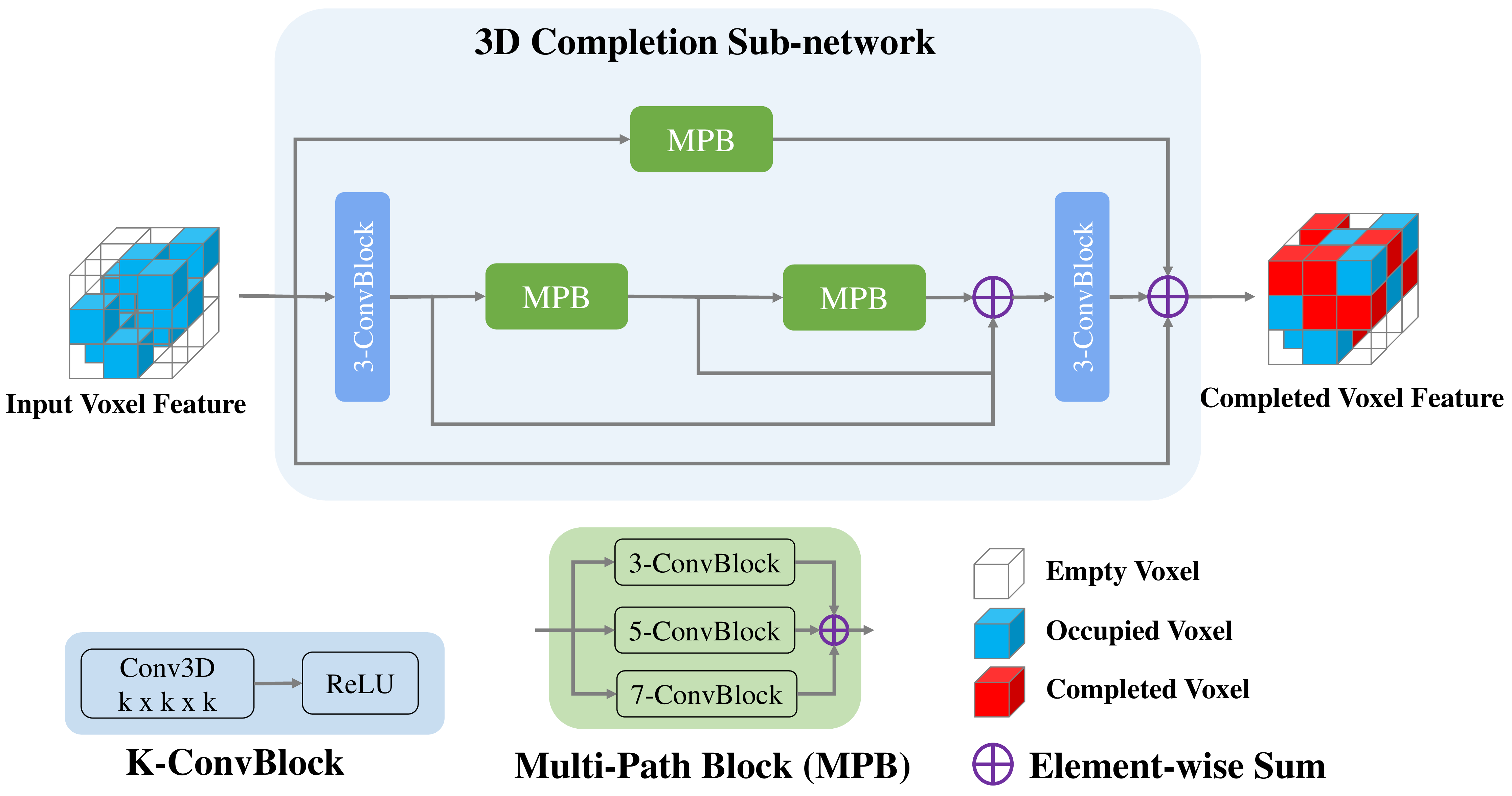}
 \vskip -0.2cm
 \caption{A schematic overview of the designed completion network. It is built upon the Multi-Path Blocks (MPBs) that contain 3$\times$3$\times$3, 5$\times$5$\times$5 and 7$\times$7$\times$7 convolution blocks (ConvBlock). In each convolution block, there is no convolution bias and no batch normalization to maintain sparsity during completion.}
 \centering
 \vskip -0.2cm
 \label{fig:network}
\end{figure}

Take JS3C-Net~\cite{js3cnet} as example. The original JS3C-Net first performs semantic segmentation and then conducts completion upon the segmentation features. Although this pipeline can benefit from the segmentation outputs, there are several drawbacks in this framework. First, the parameters in the completion sub-network are much fewer than the segmentation sub-network, thus yielding unsatisfactory completion performance. Second, there are downsampling and upsampling blocks in the completion sub-network. The downsampling operations will inevitably lose the information of the original point cloud and the upsampling operation will cause over dilation and shape distortion. %Third, the completion cost is expensive when the vanilla 3D convolution need to process all $H \times W \times L$ voxel features.

To address the aforementioned problems, we make a comprehensive overhaul of the completion sub-network. More concretely, there are three principles in the design of the completion sub-network, \ie, maintaining sparsity, no downsampling and aggregating multi-scale features.

\noindent \textbf{Maintain sparsity.} The completion sub-network needs the vanilla dense convolution for dilation while the segmentation sub-network uses sparse convolution for efficient processing. However, the bias of the convolution weight, running mean and beta of BN layers will break the sparsity of the original voxel features, thus substantially increasing the computation burden of the segmentation sub-network. Therefore, to reduce the overall computation cost and enjoy the high efficiency of sparse convolution, we remove all the convolution bias and the BN layers in the completion sub-network. In this condition, the voxel features produced by the completion sub-network can still keep the sparse property and the segmentation sub-network can use sparse convolution to process these sparse voxel features. %Since the completion sub-network only needs to process $M$ non-empty voxel features instead of all $H \times W \times L$ voxel features ($M << H \times W \times L$), the completion cost is greatly reduced. the vanilla 3D convolution is computationally intensive since it has to process the whole space. Hence, the computation burden lies in the vanilla 3D convolution.

\noindent \textbf{No downsampling.} In popular completion networks such as S3CNet~\cite{s3cnet} and JS3C-Net~\cite{js3cnet}, there are several downsampling and upsampling blocks in the completion part. The downsampling operations will inevitably lose the information of the original point cloud, causing severe completion and classification errors for small objects and crowded scenes. Therefore, we discard all downsampling and upsampling operations to relieve the information loss, maximally retaining the information of the raw point cloud. Besides, as opposed to JS3C-Net which takes the segmentation-first baseline, we adopt the completion-first principle. Concretely, we make the completion sub-network directly process the raw voxel features produced by the voxelization process. And the completion sub-network can also 

\begin{figure}[!ht]
 \centering
 \includegraphics[width=1.0\linewidth]{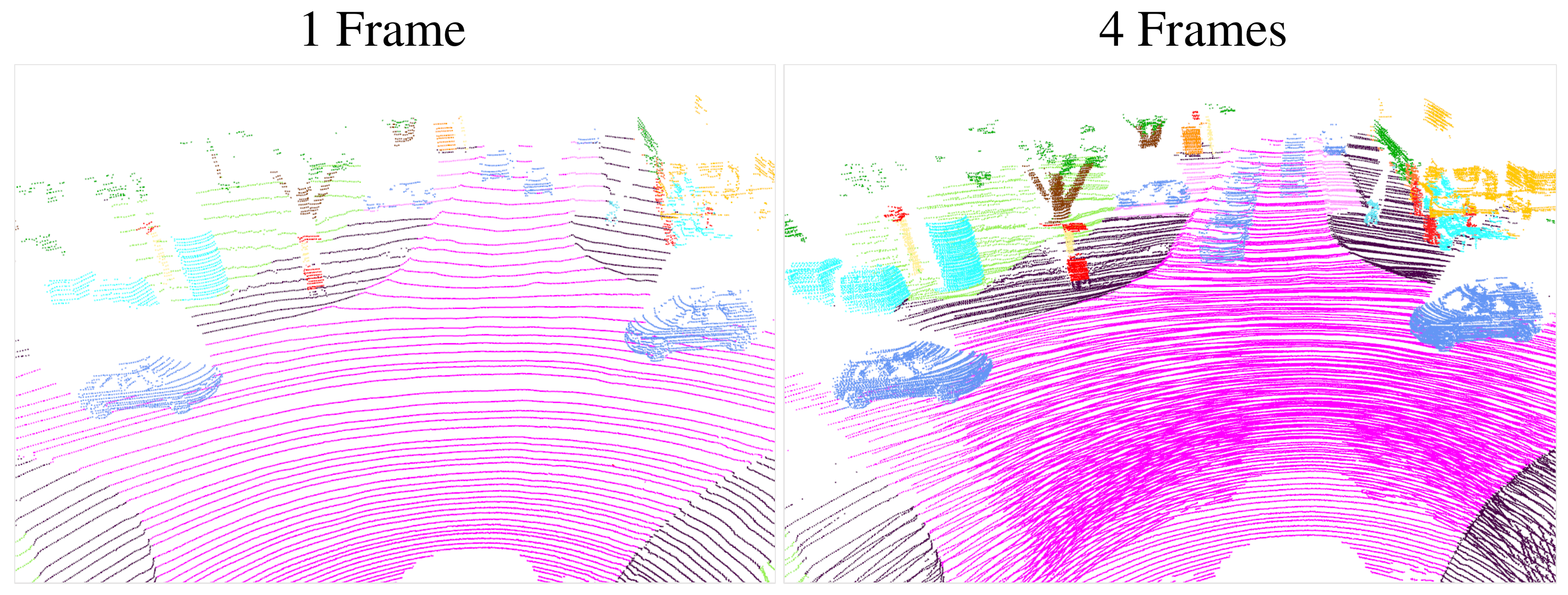}
 \vskip -0.2cm
 \caption{Comparison between single-frame and multi-frame point cloud. The multi-frame input is obviously denser than the single frame and the objects are easier to be identified, significantly reducing the completion difficulty.}
 \centering
 \vskip -0.2cm
 \label{fig:single_multi}
\end{figure}

\noindent benefit from the large number of parameters of the segmentation sub-network.

\noindent \textbf{Aggregate multi-scale features.} To aggregate multi-scale features, we design the multi-path block which is comprised of 3$\times$3$\times$3, 5$\times$5$\times$5 and 7$\times$7$\times$7 convolution blocks. As shown in Fig.~\ref{fig:network}, there are three branches in the completion sub-network. The upper branch contains one MPB and the bottom branch is a residual connection. The middle branch is constructed by a 3$\times$3$\times$3 convolution block, two MPBs and a 3$\times$3$\times$3 convolution block. After the completion sub-network, we obtain the dense completed voxel features. We extract the non-empty voxel features as well as their voxel indices from the completed voxel features. The generated sparse voxel features are sent to the segmentation sub-network to produce the voxelwise segmentation output.

\noindent \textbf{Modifications on the segmentation sub-network.} Recall that, for the segmentation part, we take the Cylinder3D~\cite{zhu2021cylindrical} as the backbone. Since the voxelwise completion labels are defined based on the cubic partition, we replace the cylindrical partition of Cylinder3D with conventional cubic partition. Besides, the original point refinement module consumes much GPU memory and brings limited gains, we discard this module to save memory usage.

% \begin{table}[!t]
% \caption{Comparison of the GPU memory usage and training speed of JS3CNet and our \algorithmname.}
% \label{completion_table}
% \centering
% \vskip -0.3cm
% \small{
% \begin{tabular}{c|c|c}
% \hline
% Network & memory usage & training speed \\
% \hline
% JS3CNet & -- & -- \\
% \hline
% \algorithmname~& -- & -- \\
% \hline
% \end{tabular}
% }
% \vspace{-2ex}
% \end{table}

\subsection{Distilling Multi-frame Knowledge}

\begin{figure}[ht]
 \centering
 \includegraphics[width=1.0\linewidth]{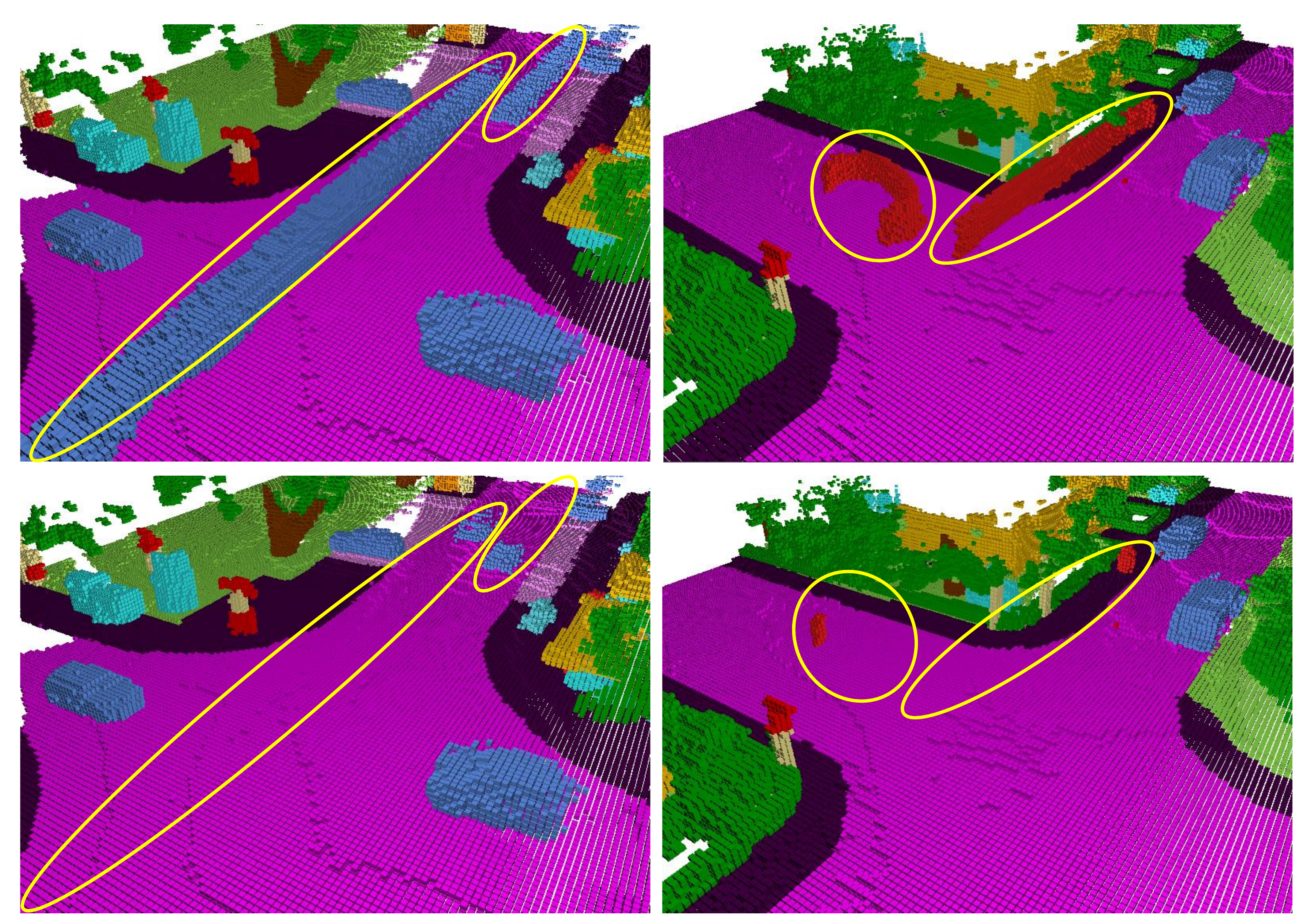}
 \vskip -0.2cm
 \caption{Top: original completion labels which have long traces of dynamic objects, \eg, car and person. Bottom: rectified completion labels. The proposed label rectification operation can effectively remove the long traces of moving objects, making the completion labels more accurate.}
 \centering
 \vskip -0.2cm
 \label{fig:trace}
\end{figure}

Since single frame point cloud is sparse and incomplete in the outdoor scenarios, directly performing semantic scene completion from the single-frame input is extremely difficult. It is natural to wonder if we can construct a multi-frame model and then distil the dense semantic knowledge from this multi-frame network. From Fig.~\ref{fig:single_multi}, it is evident that the multi-frame input significantly reduces the completion difficulty since multiple frames have much more points in the scene and objects are easier to be identified. The completion difficulty will gradually decrease as the number of input point cloud frames increases. Therefore, we construct the multi-frame teacher which takes denser point cloud as input and achieves better completion performance. 

Inspired by~\cite{pvkd2022}, we make the single-frame model distil the relation-based structural knowledge from the multi-frame teacher network. Since the original voxel features are in the sparse form, we leverage the sparse features and their indices to perform knowledge distillation. We denote the voxel features and corresponding indices of the teacher and student as $F_{T} \in \mathbb{R}^{N_{m} \times C_{f}}$, $F_{S} \in \mathbb{R}^{N_{s} \times C_{f}}$, $\mathcal{I}_{T} \in \mathbb{R}^{N_{m} \times 3}$ and $\mathcal{I}_{S} \in \mathbb{R}^{N_{s} \times 3}$, respectively. $N_{m}$ is the number of non-empty voxel features in the multi-frame, $N_{s}$ is the number of non-empty voxel features in the single-frame, $C_{f}$ is the number of channels of the voxel features. Note that the indices of teacher features and student features are sorted and $\mathcal{I}_{S}(i, j) = \mathcal{I}_{T}(i, j)$, where $i \in \{1, ..., N_{s}\}$ and $j \in \{1, 2, 3\}$. We first compute the pairwise relational knowledge of the student model:

\begin{equation}
\label{eqn:sim_voxel}
\begin{split}
\mathbf{P}_{S}(i, j) = \frac{\trans{F_{S}(i)} F_{S}(j)}{\| F_{S}(i) \|_{2} \| F_{S}(j) \|_{2}}, i, j \in \{1, ..., N_{s}\}
\end{split}
\end{equation}

The relational knowledge of the teacher model $\mathbf{P}_{T}$ is calculated in a similar way. The relational knowledge captures the similarity of each pair of voxel features and serve as important clues of the surrounding environment, which can be taken as high-level knowledge to be learned by the single-frame student model. The proposed Dense-to-Sparse Knowledge Distillation (DSKD) loss is given as below:

\begin{equation}
\label{eqn:dskd_loss}
\begin{split}
\mathcal{L}_{\mathrm{dskd}}(\mathbf{P}_{S}, \mathbf{P}_{T}) = \frac{1}{N_{s}^{2}} \sum_{i=1}^{N_{s}} \sum_{j=1}^{N_{s}} \| \mathbf{P}_{S}(i, j) - \mathbf{P}_{T}(i, j) \|_{2}^{2}.
\end{split}
\end{equation}

\subsection{Completion Label Rectification}

%##################################################################################################
\begin{algorithm}[ht]
\caption{Pseudocode of Label Rectification.}
\label{alg:code}
\algcomment{\fontsize{7.2pt}{1em}\selectfont \texttt{getInd(A, b)}: get indices of b in A; \texttt{bound}: get bound of matrix; \texttt{difference(A, B)}: difference set of A minus B.
\vspace{-2ex}
}
\definecolor{codeblue}{rgb}{0.25,0.5,0.5}
\lstset{
  backgroundcolor=\color{white},
  basicstyle=\fontsize{7.2pt}{7.2pt}\ttfamily\selectfont,
  columns=fullflexible,
  breaklines=true,
  captionpos=b,
  commentstyle=\fontsize{7.2pt}{7.2pt}\color{codeblue},
  keywordstyle=\fontsize{7.2pt}{7.2pt},
%  frame=tb,
}
\begin{lstlisting}[language=python]
# L_c: completion label (256x256x32)
# L_s: semantic segmentation label (Nx1)
# L_p: panoptic segmentation label (Nx1)
# S_c: classes of moving objects; u: unlabeled class

for C_i in S_c:
  # voxel position of completion label for class C_i
  P_c_i = getInd(L_c, C_i) 
  L_p_i = L_p[L_s == C_i] # instance label of class C_i
  P_dif = []
  if len(L_p_i) == 0:
    P_dif = P_c_i
  elif len(L_p_i) > 0 and len(P_c_i) > 0:
    # mask of voxel position bound for instance d_j
    M_c = zeros([256, 256, 32])
    # point position of class C_i
    P_i = P[L_s == C_i, :]
    # voxel position of class C_i
    P_s_i = voxelize(P_i)
    D_i = unique(L_p_i) # all instance ids
    for d_j in D_i:
        # voxel position for instance d_j
        M_j = P_s_i[L_p_i == d_j] 
        (x_min,x_max,y_min,y_max,z_min,z_max) = bound(M_j)
        M_c[x_min:x_max,y_min:y_max,z_min:z_max] = 1000
    # voxel position for instance label of class C_i
    P_p_i = getInd(M_c, 1000) 
    P_dif = difference(P_c_i, P_p_i)
  if len(P_dif) > 0:
    for p_i in range(P_dif.shape[0]):
      L_c[P_dif[p_i,0],P_dif[p_i,1],P_dif[p_i,2]] = u
\end{lstlisting}
\end{algorithm}
%\vskip{-2ex}
%##################################################################################################

\begin{table*}[!ht]
\small
\caption{Quantitative results of semantic scene completion algorithms on SemanticKITTI test set. Note that the online server still uses the \textbf{original} completion labels to evaluate algorithms. \textbf{Bold} - best in column for all single-frame methods.}
% \vskip -1.8cm
\vspace{-0.6cm}
\begin{center}
\begin{adjustbox}{width=\textwidth}
\begin{tabular}{c|c|c|ccccccccccccccccccc}
\hline
Methods & mIoU & completion & \rotatebox{90}{car} & \rotatebox{90}{bicycle} & \rotatebox{90}{motorcycle} & \rotatebox{90}{truck} & \rotatebox{90}{other-vehicle} & \rotatebox{90}{person} & \rotatebox{90}{bicyclist} & \rotatebox{90}{motorcyclist} & \rotatebox{90}{road} & \rotatebox{90}{parking} & \rotatebox{90}{sidewalk} & \rotatebox{90}{other-ground} & \rotatebox{90}{building} & \rotatebox{90}{fence} & \rotatebox{90}{vegetation} & \rotatebox{90}{trunk} & \rotatebox{90}{terrain} & \rotatebox{90}{pole} & \rotatebox{90}{traffic-sign} \\
\hline
\hline
SSCNet-Full~\cite{sscnet} & 16.1 & 50.0 & 24.3 & 0.5 & 0.8 & 1.2 & 4.3 & 0.3 & 0.3 & 0.0 & 51.2 & 27.1 & 30.8 & 6.4 & 34.5 & 19.9 & 35.3 & 18.2 & 29.0 & 13.1 & 6.7 \\ 
ESSCNet~\cite{esscnet} & 17.5 & 41.8 & 26.4 & 0.3 & 5.4 & 5.0 & 9.1 & 2.9 & 2.7 & 0.1 & 43.8 & 26.9 & 28.1 & 10.3 & 29.8 & 23.3 & 35.8 & 20.1 & 28.7 & 16.4 & 16.7 \\ 
% ESSCNet: Efficient semantic scene completion network with spatial group convolution
LMSCNet-SS~\cite{lmscnet} & 17.6 & 56.7 & 30.9 & 0.0 & 0.0 & 1.5 & 0.8 & 0.0 & 0.0 & 0.0 & 64.8 & 29.0 & 34.7 & 4.6 & 38.1 & 21.3 & 41.3 & 19.9 & 32.1 & 15.0 & 0.8 \\
TDS~\cite{tds} & 17.7 & 50.6 & 29.5 & 0.0 & 0.0 & 2.5 & 0.1 & 0.0 & 0.0 & 0.0 & 62.2 & 23.3 & 31.6 & 6.5 & 34.1 & 24.1 & 40.1 & 21.9 & 33.1 & 16.9 & 6.9 \\
% TDS: Two stream 3d semantic scene completion
UDNet~\cite{udnet} & 19.5 & \textbf{59.4} & 33.9 & 0.8 & 0.4 & 3.8 & 4.4 & 0.5 & 0.3 & 0.3 & 62.0 & 28.2 & 35.1 & 9.1 & 39.5 & 24.4 & 40.9 & 23.2 & 32.3 & 18.8 & 13.1 \\
% UDNet: Up-to-Down Network: Fusing Multi-Scale Context for 3D Semantic Scene Completion
Local-DIFs~\cite{local-dif} & 22.7 & 57.7 & 34.8 & 3.6 & 2.4 & 4.4 & 4.8 & 2.5 & 1.1 & 0.0 & 67.9 & 40.1 & 42.9 & 11.4 & 40.4 & 29.0 & 42.2 & 26.5 & 39.1 & 21.3 & 17.5\\
SSA-SC~\cite{ssa-sc} & 23.5 & 58.8 & 36.5 & 13.9 & 4.6 & 5.7 & 7.4 & 4.4 & 2.6 & 0.7 & 72.2 & 37.4 & 43.7 & 10.9 & 43.6 & 30.7 & 43.5 & 25.6 & 41.8 & 14.5 & 6.9\\
JS3C-Net~\cite{js3cnet} & 23.8 & 56.6 & 33.3 & 14.4 & 8.8 & 7.2 & 12.7 & 8.0 & 5.1 & 0.4 & 64.7 & 34.9 & 39.9 & 14.1 & 39.4 & 30.4 & 43.1 & 19.6 & 40.5 & 18.9 & 15.9\\
S3CNet~\cite{s3cnet} & 29.5 & 45.6 & 31.2 & \textbf{41.5} & \textbf{45.0} & 6.7 & 16.1 & \textbf{45.9} & \textbf{35.8} & \textbf{16.0} & 42.0 & 17.0 & 22.5 & 7.9 & \textbf{52.2} & 31.3 & 39.5 & 34.0 & 21.2 & 31.0 & 24.3\\
\hline
\algorithmname~(\#frame=1) & \textbf{36.7} & 56.1 & \textbf{46.4} & 33.2 & 34.9 & \textbf{13.8} & \textbf{29.1} & 28.2 & 24.7 & 1.8 & \textbf{68.5} & \textbf{51.3} & \textbf{49.8} & \textbf{30.7} & 38.8 & \textbf{44.7} & \textbf{46.4} & \textbf{40.1} & \textbf{48.7} & \textbf{40.4} & \textbf{25.1}\\
\algorithmname~(\#frame=4) & 47.5 & 68.5 & 59.0 & 48.5 & 51.1 & 21.0 & 37.8 & 47.6 & 35.0 & 10.5 & 79.7 & 57.7 & 60.0 & 32.8 & 50.2 & 54.1 & 56.9 & 47.1 & 58.0 & 48.3 & 48.0\\
\hline
\end{tabular}
\end{adjustbox}
\end{center}
\label{tab:ssc_semkitti}
\vspace{-0.6cm}
\end{table*}

\noindent \textbf{Generation of completion labels.} As pointed out by~\cite{completion_survey}, for outdoor semantic scene completion, the ground-truth completion labels are obtained by concatenating the segmentation labels of multiple consecutive point cloud frames. Specifically, for the $t$-th frame, the corresponding completion labels $L^{c}_{t}$ are constructed in the following way:

\begin{equation}
\label{eqn:completion_label}
\begin{split}
L^{c}_{t} = \mathrm{concat}[L^{s}_{t};T_{t+1 \rightarrow t}L^{s}_{t+1};...;T_{t+T-1 \rightarrow t}L^{s}_{t+T-1}],
\end{split}
\end{equation}
where $L^{s}_{t}$ are the segmentation labels for the $t$-th frame, $T$ is the number of frames used for concatenation, $T_{t+1 \rightarrow t}$ is the transformation matrix that transforms the coordinate from $(t+1)$-th frame to the $t$-th frame, and $\mathrm{concat}[...;...]$ is the concatenating operation. The concatenation of multiple frames will lead to long traces for those moving objects, such as car and person. A vivid example is shown in Fig.~\ref{fig:trace}. The long traces of those dynamic objects are obviously irrational and will hamper the learning of deep models.

%\noindent \textbf{Traces of dynamic objects.} We need one figure to show the smear problem of dynamic objects.

\noindent \textbf{Completion label rectification.} To remove the long traces of moving objects in the completion labels, we resort to the panoptic segmentation labels. Specifically, given the panoptic labels of class $i$, we first voxelize the labels and obtain the voxelwise panoptic labels for class $i$. For each instance of class $i$, we calculate the bound of each instance, forming a cube. We union the cubes of all instances and use them to process the original voxelwise completion labels, filtering those voxels that are outside the cubes. The process is repeated for all classes that contain dynamic objects. The detailed information of the label rectification algorithm is shown in Algorithm~\ref{alg:code}. As shown in Fig.~\ref{fig:trace}, the proposed label rectification operation can effectively remove the long traces of moving objects, making the completion labels more accurate.

\subsection{Overall objective}

The overall loss function is comprised of three terms, \ie, the cross entropy loss, the lovasz-softmax loss~\cite{berman2018the} and the proposed distillation loss.

\begin{equation}
\label{eqn:loss_func}
\begin{split}
\mathcal{L} = & \mathcal{L}_{\mathrm{ce}} + \alpha \mathcal{L}_{\mathrm{lovasz}} + \beta \mathcal{L}_{\mathrm{dskd}},
\end{split}
\end{equation}

\noindent where $\alpha$ and $\beta$ are the loss coefficients to balance the effect of each loss term.

%% file: sections/04-experiment.tex
% !TEX root = ../main.tex

\noindent \textbf{Datasets.} Following the practice of popular scene completion models~\cite{js3cnet,s3cnet}, we conduct experiments on two popular LiDAR semantic scene completion benchmarks, \ie, SemanticKITTI~\cite{behley2019semantickitti} and SemanticPOSS~\cite{semanticposs}. As to SemanticKITTI, it has 22 point cloud sequences. Sequences 00 to 10, 08 and 11 to 21 are used for training, validation and testing, respectively. 19 classes are chosen for training and evaluation after merging classes with distinct moving status and discarding classes with very few points. As for SemanticPOSS, it has 2, 988 frames and 11 classes are selected for evaluation. Although SemanticPOSS is smaller than SemanticKITTI in terms of dataset size, it is much more challenging since it contains a larger quantity of moving objects than SemanticKITTI, such as person and rider.

\noindent \textbf{Evaluation metrics.} Following~\cite{pvkd2022,zhu2021cylindrical}, we adopt the intersection-over-union (IoU) of each class and mIoU of all classes as the evaluation metric. The IoU of class $i$ is calculated via: $IoU_{i} = \frac{TP_{i}}{TP_{i}+FP_{i}+FN_{i}}$, where $TP_{i}$, $FP_{i}$ and $FN_{i}$ denote the true positive, false positive and false negative of class $i$, respectively. For semantic scene completion, the dimension of the completion label is $256 \times 256 \times 32$. We also report the completion mIoU which is the class-agnostic version of mIoU. Note that mIoU is the major evaluation metric for the semantic scene completion task.

\noindent \textbf{Implementation details.} The output size of Cylinder3D is set as $256 \times 256 \times 32$ to adapt to the completion task. The number of training epochs is set as 30 and the initial learning rate is set as 0.0015. We use Adam~\cite{kingma2015adam} as the optimizer. Gradient norm clip is set 10 to stabilize the training process. $\alpha$ and $\beta$ are set as 1 and 3, 000, respectively. We filter points that are outside the point cloud range of the voxelwise completion labels. For SemanticKITTI, we first train the model on the training set and then finetune it on both training and validation sets before submitting the predictions of test set to the online server. %We use 1 epoch to warmup the network and adopt the cosine learning rate schedule for the remaining epochs. The momentum is set 0.9 and weight decay is set 0.0001.

%\noindent \textbf{Range mismatch.} On SemanticKITTI, the range of the original point cloud is [-36.2, 36.2] m, [-36.2, 36.2] m and [-4, 2] m for x, y, z, respectively. For scene completion, the range is [0, 51.2] m, [-25.6, 25.6] m and [-2, 4.4] m for x, y, z, respectively. The range mismatch problem will cause the existence of many empty voxels, which will significantly hamper the completion performance. To address this problem, we take the intersection of two ranges and the range for the completion task is set as [0, 36.2] m, [-25.6, 25.6] m and [-2, 2] m for x, y and z, respectively.

\noindent \textbf{Baseline KD algorithms.} We take the vanilla knowledge distillation~\cite{hinton2015distilling}, FitNets~\cite{romero2015fitnets}, NST~\cite{nst}, PKT~\cite{pkt} and PVKD~\cite{pvkd2022} as baseline distillation algorithms. Vanilla knowledge distillation takes the softened logits as the distilled knowledge. FitNets directly mimics the teacher features. NST adopts the maximum mean discrepancy to minimize the distance between student features and teacher features. PKT models the teacher knowledge as a probability distribution and then forces the consistency of the probability distribution between the teacher and the student. PVKD distils the voxelwise output and inter-voxel affinity knowledge. And we discard the original pointwise output distillation and inter-point affinity distillation of PVKD since they consume much GPU memory and bring marginal gains.

\begin{table*}[ht]
\small
\caption{Quantitative results of semantic scene completion algorithms on SemanticPOSS val set. Note that the completion labels are processed by the proposed rectification algorithm. \textbf{Bold} - best in column for all single-frame methods.}
% \vskip -0.8cm
\vspace{-0.6cm}
\begin{center}
\begin{adjustbox}{width=\textwidth}
\begin{tabular}{c|c|c|ccccccccccccccccccc}
\hline
Methods & mIoU & completion & \rotatebox{90}{person} & \rotatebox{90}{rider} & \rotatebox{90}{car} & \rotatebox{90}{trunk} & \rotatebox{90}{plants} & \rotatebox{90}{traffic-sign} & \rotatebox{90}{pole} & \rotatebox{90}{building} & \rotatebox{90}{fence} & \rotatebox{90}{bike} & \rotatebox{90}{ground} \\
\hline
\hline
SSCNet-Full~\cite{sscnet} & 15.2 & 53.5 & 5.3 & 0.3 & 1.3 & 5.6 & 39.6 & 1.0 & 2.6 & 28.7 & 3.4 & 26.0 & 43.1 \\
LMSCNet-SS~\cite{lmscnet} &  16.5 & 52.3 & 7.7 & 0.3 & 0.6 & 4.0 & 37.7 & 1.9 & 8.2 & 36.8 & 13.8 & 25.8 & 45.1\\
MotionSC~\cite{wilson2022motionsc} & 17.6 & 52.7 & 7.8 & 0.5 & 0.5 & 3.9 & 39.8 & 2.2 & 8.5 & 39.2 & 13.1 & 30.8 & 47.0\\
JS3C-Net~\cite{js3cnet} & 22.7 & \textbf{58.1} & \textbf{18.9} & 0.2 & 7.1 & 3.6 & 47.8 & 2.2 & 0.0 & 46.3 & 26.6 & 43.4 & \textbf{53.2}\\
\hline
\algorithmname~(\#frame=1)& \textbf{26.3} & 56.3 & 9.2 & \textbf{3.3} & \textbf{12.4} & \textbf{11.0} & \textbf{49.6} & \textbf{3.1} & \textbf{11.1} & \textbf{50.1} & \textbf{40.8} & \textbf{48.7} & 49.8\\
\algorithmname~(\#frame=4)& 33.6 & 60.7 & 21.4 & 5.0 & 34.5 & 8.9 & 55.3 & 7.8 & 31.0 & 50.3 & 47.9 & 55.1 & 52.3\\
\hline
\end{tabular}
\end{adjustbox}
\end{center}
\label{tab:ssc_semposs}
\vspace{-0.6cm}
\end{table*}

\begin{table*}[ht]
\caption{Quantitative results of our \algorithmname~and state-of-the-art LiDAR semantic segmentation methods on SemanticKITTI test set. C3D + \algorithmname~denotes Cylinder3D initialized from the trained weight of \algorithmname. \textbf{Bold} - best in column.}
\vskip -0.2cm
\label{tab:ss_semkitti}
\centering
\begin{adjustbox}{width=\textwidth}
\begin{tabular}{c|c|c|c|c|c|c|c|c|c|c|c|c|c|c|c|c|c|c|c|c|c}
\hline
\textbf{Methods} & \textbf{mIoU} & \rotatebox{90}{Latency (ms)} & \rotatebox{90}{car} &  \rotatebox{90}{bicycle} & \rotatebox{90}{motorcycle} & \rotatebox{90}{truck} & \rotatebox{90}{other-vehicle} & \rotatebox{90}{person} & \rotatebox{90}{bicyclist} & \rotatebox{90}{motorcyclist} & \rotatebox{90}{road} & \rotatebox{90}{parking} & \rotatebox{90}{sidewalk} & \rotatebox{90}{other-ground} &
\rotatebox{90}{building} & \rotatebox{90}{fence} & \rotatebox{90}{vegetation} & \rotatebox{90}{trunk} & \rotatebox{90}{terrain} & \rotatebox{90}{pole} & \rotatebox{90}{traffic} \\
\hline
\hline
%FusionNet~\cite{zhang2020deep}  & 61.3 & -- & 95.3 & 47.5 & 37.7 & 41.8 & 34.5 & 59.5 & 56.8 & 11.9 & 91.8 & 68.8 & 77.1 & 30.8 & 92.5 & \bf{69.4} & 84.5 & 69.8 & 68.5&60.4 & 66.5 \\ 
%\hline
%KPRNet~\cite{kochanov2020kprnet}  & 63.1 & -- & 95.5&54.1& 47.9&23.6 & 42.6&65.9 & 65.0 & 16.5 & 93.2 & 73.9 & 80.6 & 30.2 & 91.7 & {68.4} & 85.7 & 69.8 & 71.2 & 58.7 & 64.1 \\
%\hline
%TORNADONet~\cite{gerdzhev2021tornado}  & 63.1 & --&94.2& 55.7& 48.1& 40.0& 38.2& 63.6& 60.1& 34.9& 89.7& 66.3& 74.5& 28.7& 91.3& 65.6& 85.6& 67.0& 71.5 & 58.0 & {65.9} \\
%\hline
SPVNAS~\cite{tang2020searching}  & 66.4 & 259 & 97.3 & 51.5 & 50.8 & 59.8 & 58.8 & 65.7 & 65.2 & 43.7 & 90.2 & 67.6 & 75.2 & 16.9 & 91.3 & 65.9 & 86.1 & 73.4 & 71.0 & 64.2 & 66.9 \\
AF2S3Net~\cite{af2s3net} & 69.7 & -- & 94.5 & 65.4 & 86.8 & 39.2 & 41.1 & \bf{80.7} & 80.4 & \bf{74.3} & 91.3 & 68.8 & 72.5 & \bf{53.5} & 87.9 & 63.2 & 70.2 & 68.5 & 53.7 & 61.5 & 71.0 \\
RPVNet~\cite{rpvnet} & 70.3 & -- & \bf{97.6} & \bf{68.4} & 68.7 & 44.2 & 61.1 & 75.9 & 74.4 & 73.4 & 93.4 & 70.3 & 80.7 & 33.3 & \bf{93.5} & 72.1 & \bf{86.5} & \bf{75.1} & 71.7 & 64.8 & 61.4 \\
PVKD~\cite{pvkd2022} & 71.2 & \bf{76} & 97.0 & 67.9 & \bf{69.3} & 53.5 & 60.2 & 75.1 & 73.5 & 50.5 & 91.8 & 70.9 & 77.5 & 41.0 & 92.4 & 69.4 & \bf{86.5} & 73.8 & \bf{71.9} & 64.9 & 65.8 \\
%\hline
2DPASS~\cite{yan20222dpass} & \bf{72.9} & -- & 97.0 & 63.6 & 63.4 & \bf{61.1} & 61.5 & 77.9 & \bf{81.3} & 74.1 & 89.7 & 67.4 & 74.7 & 40.0 & 93.5 & \bf{72.9} & 86.2 & 73.9 & 71.0 & 65.0 & 70.4 \\
\hline
Cylinder3D~\cite{zhu2021cylindrical}  & 68.9 & \multirow{2}*{170} & 97.1 & 67.6 & 63.8 & 50.8 & 58.5 & 73.7 & 69.2 & 48.0 & 92.2 & 65.0 & 77.0 & 32.3 & 90.7 & 66.5 & 85.6 & 72.5 & 69.8 & 62.4 & 66.2 \\
C3D + \algorithmname~& 71.5 & ~ & 97.5 & 60.9 & 56.3 & 58.6 & \bf{65.9} & 70.7 & 71.8 & 58.7 & \bf{93.6} & \bf{72.1} & \bf{80.9} & 36.2 & 93.3 & 72.1 & 86.2 & 74.1 & 71.6 & \bf{66.7} & \bf{71.8} \\
\hline
\end{tabular}
\end{adjustbox}
\vspace{-2ex}
\end{table*}

\subsection{Results}

\noindent \textbf{Quantitative results.} We summarize the performance of \algorithmname~and state-of-the-art semantic scene completion methods in Table~\ref{tab:ssc_semkitti} and~\ref{tab:ssc_semposs}. On SemanticKITTI, our \algorithmname~significantly outperforms other scene completion algorithms in terms of mIoU. For example, our \algorithmname~is \textbf{7.2} mIoU higher than S3CNet~\cite{s3cnet}. On classes such as car, other-vehicle, road, parking, sidewalk, fense, terrain and other-ground, the performance gap between \algorithmname~and S3CNet is more than \textbf{10} IoU. Our \algorithmname~also achieves superior performance on SemanticPOSS val set. On classes such as car, trunk, pole, fence and bike, \algorithmname~is at least \textbf{5} IoU higher than JS3C-Net~\cite{js3cnet}. The impressive performance on two large-scale benchmarks strongly demonstrate the superiority of our \algorithmname~.

Besides, we use the trained weight of the segmentation sub-network as initialization to train Cylinder3D on the SemanticKITTI semantic segmentation task. From Table~\ref{tab:ss_semkitti}, Cylinder3D initialized from trained weight of the completion task outperforms the original Cylinder3D model by \textbf{2.6} mIoU, and achieves impressive segmentation performance among various competitive LiDAR segmentation models such as 2DPASS~\cite{yan20222dpass}, PVKD~\cite{pvkd2022} and RPVNet~\cite{rpvnet}. The encouraging results show that knowledge learned in the completion task is also beneficial to the segmentation task.

\begin{figure*}[t]
 \centering
 \includegraphics[width=1.0\linewidth]{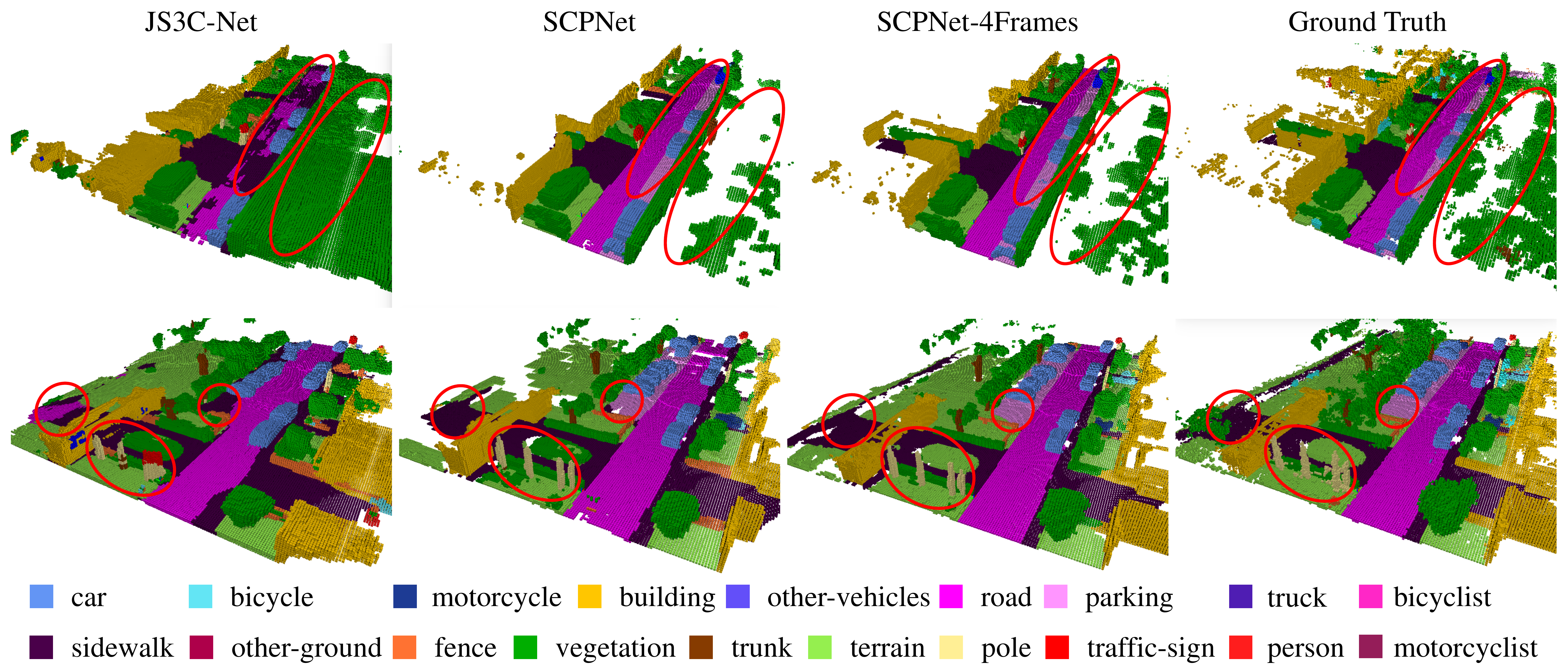}
 \vskip -0.1cm
 \caption{Visual comparison of different methods on the SemanticKITTI validation set. From left to right: predictions of JS3C-Net~\cite{js3cnet}, \algorithmname~(single frame), \algorithmname~(multi-frame) and ground-truth. Regions that have large prediction errors are highlighted by red ellipses.}
 \centering
 \vskip -0.2cm
 \label{fig:visual_compare}
\end{figure*}

\begin{table}[!t]
\caption{Comparison between different knowledge distillation algorithms and the proposed DSKD on the SemanticKITTI val set.}
\label{tab:kd_com}
\centering
\vskip -0.1cm
%\small{
\begin{tabular}{l|c}
\hline
Methods & mIoU \\ %& Completion \\
\hline
\algorithmname~w/o DSKD & 34.4 \\ %& 48.5 \\
\hline
+ KD~\cite{hinton2015distilling} & 33.8 \\ %& -- \\
+ FitNets~\cite{romero2015fitnets} & 33.8 \\ %& 48.8 \\
+ PKT~\cite{pkt} & 34.6 \\ %& 50.2 \\
+ PVKD~\cite{pvkd2022} & 36.2 \\ %& 49.3 \\
+ NST~\cite{nst} & 36.3 \\ %& 48.5 \\
\hline
+ DSKD & \bf{37.2} \\ %& \bf{49.9} \\
\hline
\end{tabular}
\vspace{-2ex}
\end{table}

\noindent \textbf{Comparison with baseline KD algorithms.} From Table~\ref{tab:kd_com}, it is evident that the proposed DSKD method can bring more gains than conventional knowledge distillation algorithms. For instance, compared with FitNets~\cite{romero2015fitnets} which directly mimics the teacher features, our DSKD can bring more than \textbf{2} mIoU, showing the effectiveness of the proposed relation-based distillation algorithm. The vanilla KD objective and FitNets hamper the performance of the base model, indicating that directly mimicking the logits or features can not boost the completion performance.

\noindent \textbf{Qualitative results.} We also provide visual comparison between JS3C-Net~\cite{js3cnet}, \algorithmname~(single-frame) and \algorithmname~(multi-frame). As can be seen from Fig.~\ref{fig:visual_compare}, our \algorithmname~(single-frame) make more accurate completion predictions than JS3C-Net on road and vegetation. On long, thin objects such as poles, our single-frame model also yields high-quality completion results compared with JS3C-Net. The predictions of our single-frame model also resemble those of the multi-frame network, demonstrating the efficacy of the proposed DSKD algorithm.
%In addition to quantitative comparison,

\subsection{Ablation studies}

%We perform ablation studies to verify the effect of the completion sub-network, DSKD, completion label rectification and the downsampling operation on the final performance. 
Experiments are conducted in SemanticKITTI val set.

\begin{table*}[!t]
\caption{Impact of completion label rectification on the performance.}
\label{tab:label_refine}
\centering
\vskip -0.2cm
%\small{
\begin{adjustbox}{width=\textwidth}
\begin{tabular}{c|c|c|ccccccccccccccccccc}
\hline
Methods & mIoU & completion & \rotatebox{90}{car} & \rotatebox{90}{bicycle} & \rotatebox{90}{motorcycle} & \rotatebox{90}{truck} & \rotatebox{90}{other-vehicle} & \rotatebox{90}{person} & \rotatebox{90}{bicyclist} & \rotatebox{90}{motorcyclist} & \rotatebox{90}{road} & \rotatebox{90}{parking} & \rotatebox{90}{sidewalk} & \rotatebox{90}{other-ground} & \rotatebox{90}{building} & \rotatebox{90}{fence} & \rotatebox{90}{vegetation} & \rotatebox{90}{trunk} & \rotatebox{90}{terrain} & \rotatebox{90}{pole} & \rotatebox{90}{traffic-sign} \\
\hline
\hline
%Baseline & 34.4 & 48.5 & 48.5 & 26.4 & 28.1 & 54.6 & \textbf{41.7} & 14.5 & 13.1 & 0.0 & 70.2 & \textbf{58.3} & \textbf{51.3} & \textbf{2.9} & 31.7 & 30.4 & 37.9 & \textbf{31.6} & 49.2 & 36.7 & 25.7 \\
%Baseline + label rectification & \textbf{37.6} & \textbf{49.9} & \textbf{56.4} & \textbf{27.1} & \textbf{31.1} & \textbf{58.4} & 40.7 & \textbf{38.6} & \textbf{33.3} & \textbf{0.7} & \textbf{70.5} & 57.5 & 51.0 & 2.0 & \textbf{33.8} & 30.4 & \textbf{38.7} & 30.7 & \textbf{50.4} & \textbf{37.6} & \textbf{26.2} \\
\algorithmname~ & 37.2 & 49.9 & 50.5 & \textbf{28.5} & 31.7 & \textbf{58.4} & 41.4 & 19.4 & 19.9 & 0.2 & 70.5 & \textbf{60.9} & 52.0 & \textbf{20.2} & \textbf{34.1} & 33.0 & 35.3 & \textbf{33.7} & 51.9 & 38.3 & 27.5 \\
\algorithmname~+ label rectification & \textbf{40.8} & \textbf{52.1} & \textbf{58.6} & 27.0 & \textbf{34.7} & 54.2 & \textbf{41.9} & \textbf{43.9} & \textbf{46.0} & \textbf{17.8} & \textbf{70.6} & 60.2 & \textbf{53.1} & 7.7 & 33.9 & 32.2 & \textbf{41.9} & 32.0 & \textbf{52.4} & \textbf{38.7} & \textbf{28.9} \\
\hline
\end{tabular}
\end{adjustbox}
\vspace{-1ex}
\end{table*}

\noindent \textbf{Effect of completion label rectification.} We report the performance of our \algorithmname~on completion labels with and without rectification. From Table~\ref{tab:label_refine}, it is evident that the proposed rectification strategy greatly enhances the performance of \algorithmname~on those dynamic objects, \eg, car and person. For example, the completion label rectification can bring \textbf{8.1}, \textbf{24.5},  \textbf{26.1} and \textbf{17.6} IoU improvement on car, person, bicyclist and motorcyclist, respectively. The impressive performance gains strongly demonstrate the effectiveness of the label rectification algorithm.

\begin{table}[!t]
\caption{Impact of the completion sub-network, DSKD and downsampling operation on the performance.}
\label{tab:dskd}
\centering
\vskip -0.3cm
%\small{
\subfloat[Completion sub-network]{\begin{tabular}{c|c}
\hline
& mIoU \\
\cline{1-2}
JS3C-Net~\cite{js3cnet} & 24.0 \\ %& 56.6 \\
JS3C-Net + completion-subnet & \bf{30.8} \\ %& -- \\
\hline
\end{tabular}}\qquad \qquad \qquad
\subfloat[DSKD]{\begin{tabular}{c|c}
\hline
& mIoU \\
\cline{1-2}
\algorithmname~w/o DSKD & 34.4 \\ %& 48.5 \\
\algorithmname~w/ DSKD & \bf{37.2} \\ %& \textbf{49.9} \\
\hline
\end{tabular}}\qquad \qquad \qquad
\subfloat[Downsampling operation]{\begin{tabular}{c|c}
\hline
& mIoU \\
\cline{1-2}
\algorithmname~w/o downsampling & \bf{37.2} \\ %& 49.9 \\
\algorithmname~w/ downsampling & 33.1 \\ %& 51.0 \\
\hline
\end{tabular}}
%}
\vspace{-2ex}
\end{table}

\noindent \textbf{Effect of the completion sub-network.} To examine the effect of our completion sub-network, we add it to JS3C-Net. The detailed performance is shown in Table~\ref{tab:dskd} (a). Our completion sub-network can bring \textbf{6.8} mIoU improvement to JS3C-Net, which strongly demonstrates the effectiveness and generalization of the proposed completion sub-network.

\noindent \textbf{Effect of DSKD.} We compare the performance of our \algorithmname~with and without the proposed DSKD in Table~\ref{tab:dskd} (b). The proposed distillation method can bring \textbf{2.8} mIoU improvement to our \algorithmname~, showing the benefit of distilling relation-based knowledge from the multi-frame model.

\noindent \textbf{Effect of the downsampling operation.} We add the downsampling operation to our completion sub-network and examine its effect. As reported in Table~\ref{tab:dskd} (c), the downsampling operation hampers the completion performance of our \algorithmname~. Specifically, the completion performance of \algorithmname~decreases from 37.2 mIoU to 33.1 mIoU. The negative results show that the no-downsampling principle is vital to the success of the completion sub-network redesign.

%% file: main.bbl
\begin{thebibliography}{10}\itemsep=-1pt

\bibitem{behley2019semantickitti}
Jens Behley, Martin Garbade, Andres Milioto, Jan Quenzel, Sven Behnke, Cyrill
  Stachniss, and Jurgen Gall.
\newblock Semantic{KITTI}: A {D}ataset for {S}emantic {S}cene {U}nderstanding
  of {L}idar {S}equences.
\newblock In {\em IEEE International Conference on Computer Vision}, pages
  9297--9307, 2019.

\bibitem{berman2018the}
Maxim {Berman}, Amal~Rannen {Triki}, and Matthew~B. {Blaschko}.
\newblock The {L}ovasz-{S}oftmax {L}oss: A {T}ractable {S}urrogate for the
  {O}ptimization of the {I}ntersection-{O}ver-{U}nion {M}easure in {N}eural
  {N}etworks.
\newblock In {\em IEEE Conference on Computer Vision and Pattern Recognition},
  volume 2018, pages 4413--4421, 2018.

\bibitem{s3cnet}
Ran Cheng, Christopher Agia, Yuan Ren, Xinhai Li, and Liu Bingbing.
\newblock S3{CN}et: {A} {S}parse {S}emantic {S}cene {C}ompletion {N}etwork for
  {L}i{DAR} {P}oint {C}louds.
\newblock In {\em Conference on Robot Learning}, pages 2148--2161. PMLR, 2021.

\bibitem{af2s3net}
Ran Cheng, Ryan Razani, Ehsan Taghavi, Enxu Li, and Bingbing Liu.
\newblock (af)2-{S3N}et: Attentive {F}eature {F}usion with {A}daptive {F}eature
  {S}election for {S}parse {S}emantic {S}egmentation {N}etwork.
\newblock In {\em IEEE Conference on Computer Vision and Pattern Recognition},
  pages 12547--12556, 2021.

\bibitem{scancomplete}
Angela Dai, Daniel Ritchie, Martin Bokeloh, Scott Reed, J{\"u}rgen Sturm, and
  Matthias Nie{\ss}ner.
\newblock Scancomplete: {L}arge-scale {S}cene {C}ompletion and {S}emantic
  {S}egmentation for 3{D} {S}cans.
\newblock In {\em IEEE Conference on Computer Vision and Pattern Recognition},
  pages 4578--4587, 2018.

\bibitem{survey_point_cloud_completion}
Ben Fei, Weidong Yang, Wenming Chen, Zhijun Li, Yikang Li, Tao Ma, Xing Hu, and
  Lipeng Ma.
\newblock Comprehensive {R}eview of {D}eep {L}earning-based 3{D} {P}oint
  {C}louds {C}ompletion {P}rocessing and {A}nalysis.
\newblock {\em arXiv preprint arXiv:2203.03311}, 2022.

\bibitem{tds}
Martin Garbade, Yueh-Tung Chen, Johann Sawatzky, and Juergen Gall.
\newblock Two {S}tream 3{D} {S}emantic {S}cene {C}ompletion.
\newblock In {\em IEEE Conference on Computer Vision and Pattern Recognition
  Workshops}, pages 0--0, 2019.

\bibitem{hinton2015distilling}
Geoffrey Hinton, Oriol Vinyals, and Jeff Dean.
\newblock Distilling the {K}nowledge in a {N}eural {N}etwork.
\newblock {\em Statistics}, 1050:9, 2015.

\bibitem{hou2020inter}
Yuenan Hou, Zheng Ma, Chunxiao Liu, Tak-Wai Hui, and Chen~Change Loy.
\newblock Inter-{R}egion {A}ffinity {D}istillation for {R}oad {M}arking
  {S}egmentation.
\newblock In {\em IEEE Conference on Computer Vision and Pattern Recognition},
  pages 12486--12495, 2020.

\bibitem{sad}
Yuenan Hou, Zheng Ma, Chunxiao Liu, and Chen~Change Loy.
\newblock Learning {L}ightweight {L}ane {D}etection {CNN}s by {S}elf
  {A}ttention {D}istillation.
\newblock In {\em IEEE International Conference on Computer Vision}, pages
  1013--1021, 2019.

\bibitem{fmnet}
Yuenan Hou, Zheng Ma, Chunxiao Liu, and Chen~Change Loy.
\newblock Learning to {S}teer by {M}imicking {F}eatures from {H}eterogeneous
  {A}uxiliary {N}etworks.
\newblock In {\em Association for the Advancement of Artificial Intelligence},
  volume~33, pages 8433--8440, 2019.

\bibitem{pvkd2022}
Yuenan Hou, Xinge Zhu, Yuexin Ma, Chen~Change Loy, and Yikang Li.
\newblock Point-to-{V}oxel {K}nowledge {D}istillation for {L}i{DAR} {S}emantic
  {S}egmentation.
\newblock In {\em IEEE Conference on Computer Vision and Pattern Recognition},
  pages 8479--8488, 2022.

\bibitem{nst}
Zehao Huang and Naiyan Wang.
\newblock Like {W}hat {Y}ou {L}ike: {K}nowledge {D}istill via {N}euron
  {S}electivity {T}ransfer.
\newblock {\em arXiv preprint arXiv:1707.01219}, 2017.

\bibitem{kingma2015adam}
Diederik~P Kingma and Jimmy Ba.
\newblock Adam: A {M}ethod for {S}tochastic {O}ptimization.
\newblock In {\em International Conference on Learning Representations}, 2015.

\bibitem{satnet}
Shice Liu, Yu Hu, Yiming Zeng, Qiankun Tang, Beibei Jin, Yinhe Han, and Xiaowei
  Li.
\newblock See and {T}hink: {D}isentangling {S}emantic {S}cene {C}ompletion.
\newblock {\em Advances in Neural Information Processing Systems}, 31, 2018.

\bibitem{semanticposs}
Yancheng Pan, Biao Gao, Jilin Mei, Sibo Geng, Chengkun Li, and Huijing Zhao.
\newblock Semantic{POSS}: {A} {P}oint {C}loud {D}ataset with {L}arge {Q}uantity
  of {D}ynamic {I}nstances.
\newblock In {\em IEEE Intelligent Vehicles Symposium}, pages 687--693. IEEE,
  2020.

\bibitem{pkt}
Nikolaos Passalis and Anastasios Tefas.
\newblock Learning {D}eep {R}epresentations with {P}robabilistic {K}nowledge
  {T}ransfer.
\newblock In {\em European Conference on Computer Vision}, pages 268--284,
  2018.

\bibitem{local-dif}
Christoph~B Rist, David Emmerichs, Markus Enzweiler, and Dariu~M Gavrila.
\newblock Semantic {S}cene {C}ompletion using {L}ocal {D}eep {I}mplicit
  {F}unctions on {L}i{DAR} {D}ata.
\newblock {\em IEEE Transactions on Pattern Analysis and Machine Intelligence},
  44(10):7205--7218, 2021.

\bibitem{lmscnet}
Luis Roldao, Raoul de Charette, and Anne Verroust-Blondet.
\newblock L{MSCN}et: {L}ightweight {M}ultiscale 3{D} {S}emantic {C}ompletion.
\newblock In {\em International Conference on 3D Vision (3DV)}, pages 111--119.
  IEEE, 2020.

\bibitem{completion_survey}
Luis Roldao, Raoul De~Charette, and Anne Verroust-Blondet.
\newblock 3d {S}emantic {S}cene {C}ompletion: {A} {S}urvey.
\newblock {\em International Journal of Computer Vision}, pages 1--28, 2022.

\bibitem{romero2015fitnets}
Adriana {Romero}, Nicolas {Ballas}, Samira~Ebrahimi {Kahou}, Antoine
  {Chassang}, Carlo {Gatta}, and Yoshua {Bengio}.
\newblock Fit{N}ets: Hints for {T}hin {D}eep {N}ets.
\newblock In {\em International Conference on Learning Representations}, 2015.

\bibitem{sscnet}
Shuran Song, Fisher Yu, Andy Zeng, Angel~X Chang, Manolis Savva, and Thomas
  Funkhouser.
\newblock Semantic {S}cene {C}ompletion from a {S}ingle {D}epth {I}mage.
\newblock In {\em IEEE Conference on Computer Vision and Pattern Recognition},
  pages 1746--1754, 2017.

\bibitem{tang2020searching}
Haotian Tang, Zhijian Liu, Shengyu Zhao, Yujun Lin, Ji Lin, Hanrui Wang, and
  Song Han.
\newblock Searching {E}fficient 3{D} {A}rchitectures with {S}parse
  {P}oint-{V}oxel {C}onvolution.
\newblock In {\em European Conference on Computer Vision}, pages 685--702.
  Springer, 2020.

\bibitem{tung2019similarity}
Fred {Tung} and Greg {Mori}.
\newblock Similarity-{P}reserving {K}nowledge {D}istillation.
\newblock In {\em IEEE International Conference on Computer Vision}, pages
  1365--1374, 2019.

\bibitem{wilson2022motionsc}
Joey Wilson, Jingyu Song, Yuewei Fu, Arthur Zhang, Andrew Capodieci, Paramsothy
  Jayakumar, Kira Barton, and Maani Ghaffari.
\newblock Motion{SC}: {D}ata {S}et and {N}etwork for {R}eal-{T}ime {S}emantic
  {M}apping in {D}ynamic {E}nvironments.
\newblock {\em arXiv preprint arXiv:2203.07060}, 2022.

\bibitem{xing2021categorical}
Xiaohan Xing, Yuenan Hou, Hang Li, Yixuan Yuan, Hongsheng Li, and Max Q-H Meng.
\newblock Categorical {R}elation-preserving {C}ontrastive {K}nowledge
  {D}istillation for {M}edical {I}mage {C}lassification.
\newblock In {\em Medical Image Computing and Computer Assisted Intervention},
  pages 163--173. Springer, 2021.

\bibitem{xu2022mind}
Guodong Xu, Yuenan Hou, Ziwei Liu, and Chen~Change Loy.
\newblock Mind the {G}ap in {D}istilling {S}tyle{GAN}s.
\newblock In {\em European Conference on Computer Vision}, pages 423--439.
  Springer, 2022.

\bibitem{rpvnet}
Jianyun Xu, Ruixiang Zhang, Jian Dou, Yushi Zhu, Jie Sun, and Shiliang Pu.
\newblock {RPV}net: A {D}eep and {E}fficient {R}ange-{P}oint-{V}oxel {F}usion
  {N}etwork for {L}idar {P}oint {C}loud {S}egmentation.
\newblock In {\em IEEE International Conference on Computer Vision}, pages
  16024--16033, October 2021.

\bibitem{js3cnet}
Xu Yan, Jiantao Gao, Jie Li, Ruimao Zhang, Zhen Li, Rui Huang, and Shuguang
  Cui.
\newblock Sparse {S}ingle {S}weep {L}i{DAR} {P}oint {C}loud {S}egmentation via
  {L}earning {C}ontextual {S}hape {P}riors from {S}cene {C}ompletion.
\newblock In {\em Proceedings of the AAAI Conference on Artificial
  Intelligence}, volume~35, pages 3101--3109, 2021.

\bibitem{yan20222dpass}
Xu Yan, Jiantao Gao, Chaoda Zheng, Chao Zheng, Ruimao Zhang, Shuguang Cui, and
  Zhen Li.
\newblock 2{DPASS}: 2{D} {P}riors {A}ssisted {S}emantic {S}egmentation on
  {L}i{DAR} {P}oint {C}louds.
\newblock In {\em European Conference on Computer Vision (ECCV)}, 2022.

\bibitem{sparsekd}
Jihan Yang, Shaoshuai Shi, Runyu Ding, Zhe Wang, and Xiaojuan Qi.
\newblock Towards {E}fficient 3{D} {O}bject {D}etection with {K}nowledge
  {D}istillation.
\newblock {\em arXiv preprint arXiv:2205.15156}, 2022.

\bibitem{ssa-sc}
Xuemeng Yang, Hao Zou, Xin Kong, Tianxin Huang, Yong Liu, Wanlong Li, Feng Wen,
  and Hongbo Zhang.
\newblock Semantic {S}egmentation-assisted {S}cene {C}ompletion for {L}i{DAR}
  {P}oint {C}louds.
\newblock In {\em IEEE International Conference on Intelligent Robots and
  Systems}, pages 3555--3562. IEEE, 2021.

\bibitem{zagoruyko2016paying}
Sergey Zagoruyko and Nikos Komodakis.
\newblock Paying {M}ore {A}ttention to {A}ttention: Improving the {P}erformance
  of {C}onvolutional {N}eural {N}etworks via {A}ttention {T}ransfer.
\newblock In {\em International Conference on Learning Representations}, 2017.

\bibitem{esscnet}
Jiahui Zhang, Hao Zhao, Anbang Yao, Yurong Chen, Li Zhang, and Hongen Liao.
\newblock Efficient semantic scene completion network with spatial group
  convolution.
\newblock In {\em European Conference on Computer Vision}, pages 733--749,
  2018.

\bibitem{zhu2021cylindrical-tpami}
Xinge Zhu, Hui Zhou, Tai Wang, Fangzhou Hong, Wei Li, Yuexin Ma, Hongsheng Li,
  Ruigang Yang, and Dahua Lin.
\newblock Cylindrical and {A}symmetrical 3{D} {C}onvolution {N}etworks for
  {L}idar-based {P}erception.
\newblock {\em IEEE Transactions on Pattern Analysis and Machine Intelligence},
  2021.

\bibitem{zhu2021cylindrical}
Xinge Zhu, Hui Zhou, Tai Wang, Fangzhou Hong, Yuexin Ma, Wei Li, Hongsheng Li,
  and Dahua Lin.
\newblock Cylindrical and {A}symmetrical 3{D} {C}onvolution {N}etworks for
  {L}idar {S}egmentation.
\newblock In {\em IEEE Conference on Computer Vision and Pattern Recognition},
  pages 9939--9948, 2021.

\bibitem{udnet}
Hao Zou, Xuemeng Yang, Tianxin Huang, Chujuan Zhang, Yong Liu, Wanlong Li, Feng
  Wen, and Hongbo Zhang.
\newblock Up-to-{D}own {N}etwork: {F}using {M}ulti-{S}cale {C}ontext for 3{D}
  {S}emantic {S}cene {C}ompletion.
\newblock In {\em IEEE/RSJ International Conference on Intelligent Robots and
  Systems}, pages 16--23. IEEE, 2021.

\end{thebibliography}
